\documentclass[sigconf, 10pt]{acmart}
\renewcommand\footnotetextcopyrightpermission[1]{} 

\def\BibTeX{{\rm B\kern-.05em{\sc i\kern-.025em b}\kern-.08emT\kern-.1667em\lower.7ex\hbox{E}\kern-.125emX}}

\AtBeginDocument{%
  \providecommand\BibTeX{{%
    Bib\TeX}}}

\setcopyright{acmlicensed}
\copyrightyear{2024}
\acmYear{2024}
\acmDOI{XXXXXXX.XXXXXXX}

\acmConference[{SenSys'24}]{ACM Conference on Embedded Networked Sensor Systems}{November 2024}{Hangzhou, China}
\acmISBN{978-1-4503-XXXX-X/18/06}

\usepackage[ruled,linesnumbered]{algorithm2e}
\usepackage{subcaption}
\usepackage{adjustbox,xspace}

\begin{document}

%####################################################################
%——————————————————标题—————————————————
%####################################################################
\title{Aucamp: An Underwater Camera-Based Multi-Robot Platform with Low-Cost, Distributed, and Robust Localization
}

\title[Aucamp]{\LARGE{Aucamp: An Underwater Camera-Based Multi-Robot Platform \\ with Low-Cost, Distributed, and Robust Localization}}

\author{Jisheng Xu}
\email{Jimmy\_xu@sjtu.edu.cn}
%%\orcid{1234-5678-9012}
\affiliation{%
  \institution{Shanghai Jiao Tong University}
  %\city{the Department of Automation}
%%  \state{Ohio}
  \country{China}
%  \postcode{200000}
}
\author{Ding Lin}
\email{tysjdlin@sjtu.edu.cn}
%%\orcid{1234-5678-9012}
\affiliation{%
  \institution{Shanghai Jiao Tong University}
  %\city{the Department of Automation}
%%  \state{Ohio}
  \country{China}
%  \postcode{200000}
}
\author{Pangkit Fong}
\email{fpjgaoge@sjtu.edu.cn}
\affiliation{%
  \institution{Shanghai Jiao Tong University}
  %\city{the Department of Automation}
%%  \state{Ohio}
  \country{China}
%  \postcode{200000}
}

\author{Chongrong Fang}
\email{crfang@sjtu.edu.cn}
\affiliation{%
  \institution{Shanghai Jiao Tong University}
  %\city{the Department of Automation}
%%  \state{Ohio}
  \country{China}
%  \postcode{200000}
}
\author{Jianping He}
\email{jphe@sjtu.edu.cn}
%%\orcid{1234-5678-9012}
\affiliation{%
  \institution{Shanghai Jiao Tong University}
  %\city{the Department of Automation}
%%  \state{Ohio}
  \country{China}
%  \postcode{200000}
}

\author{Xiaoming Duan}
\email{xduan@sjtu.edu.cn}
%%\orcid{1234-5678-9012}
\affiliation{%
  \institution{Shanghai Jiao Tong University}
  %\city{the Department of Automation}
%%  \state{Ohio}
  \country{China}
%  \postcode{200000}
}

%####################################################################
%——————————————————摘要—————————————————
%####################################################################
\begin{abstract}
%本文提出了Aucamp，一个面向分布式定位的水下多机器人平台，采用了低成本的单目相机感知。
This paper introduces an underwater multi-robot platform, named Aucamp, characterized by cost-effective monocular-camera-based sensing, distributed protocol and robust orientation control for localization. 
%With each robot carrying a single low-cost monocular camera (hundred times cheaper than the sensors commonly used for localization), w
We utilize the clarity feature to measure the distance, 
%employ lightweight underwater image enhancement algorithm to improve imaging performance, 
present the monocular imaging model, and estimate the position of the target object. 
We achieve global positioning in our platform by designing a distributed update protocol. The distributed algorithm enables the perception process to simultaneously cover a broader range,  and greatly improves the accuracy and robustness of the positioning.
%, thereby reducing the relative localization error of our platform to less than 0.4\%.
Moreover, the explicit dynamics model of the robot in our platform is obtained, based on which, we propose a robust orientation control framework. The control system ensures that the platform maintains a balanced posture for each robot, thereby ensuring the stability of the localization system. The platform can swiftly recover from an forced unstable state to a stable horizontal posture.
%(within about 2 seconds).
%Additionally, we conduct extensive experiments to evaluate the performance of our platform across various aspects, including the depth imaging efficacy, the control robustness, and the distributed localization accuracy. 
%Besides, we present some application scenarios of our platform, including underwater creature tracking and distributed underwater 3D scene reconstruction.
Additionally, we conduct extensive experiments and application scenarios to evaluate the performance of our platform.
The proposed new platform may provide support for extensive marine exploration by underwater sensor networks.
% 意义？
\end{abstract}


%% A "teaser" image appears between the author and affiliation information and the body of the document, and typically spans the page.
\begin{teaserfigure}
  \centering\includegraphics[width=0.9\textwidth]{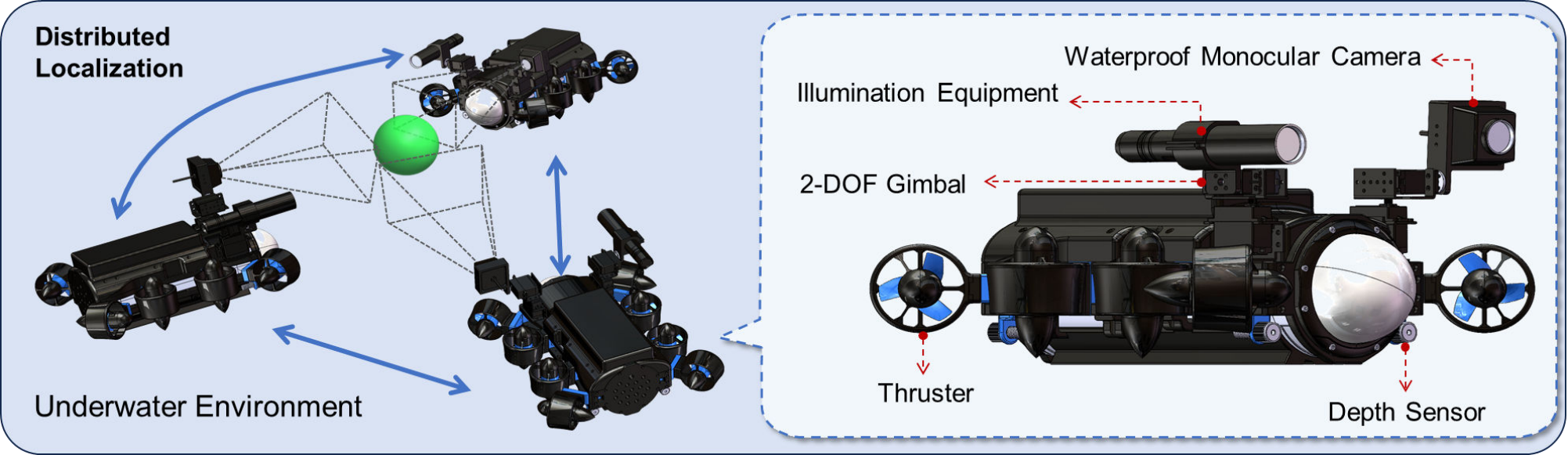}
  \caption{Aucamp is a underwater multi-robot platform that provides low-cost and robust distributed localization.}
  \Description{The overview of the whole platform.}
  \label{fig:equipment}
\end{teaserfigure}


\maketitle

%####################################################################
%——————————————————intro—————————————————
%####################################################################
% 原本的讲故事思路：水下机器人集群很多应用，他们的共性问题是水下感知，水下感知难点在于传感器受限。水下能用的有三种传感器，基于相机的传感器最为经济，但仍需要解决一些问题，其一是固定的补光成像配置，其二是视觉定位能力的缺失。本文提出了一种新的平台……

%新的讲故事思路：水下机器人很多应用，本文提出一种新的水下机器人平台，要求三点：低成本的水下感知、分布式的水下定位、鲁棒的水下控制。如果实现这样的设计，能够使得某些事情的实现称为可能。而要实现这样一个平台，可能涉及诸多挑战。首先是传感器受限（有三种传感器，其他都贵，而基于相机的难以测距。本文提出了基于单目相机的感知测距算法），其次是缺少水下的分布式定位方式（地面上可以进行GPS、UWB进行全局定位，但水下全是陌生环境，无法获得全局定位，需要有综合局部定位来实现全局定位的算法。本文采用了基于一致性的定位算法，进而能够实现最优补光，辅助全局定位），最后是水下机器人需要能够在洋流仍保证稳定的姿态（启发自nature，水压计可以对水下环境进行一定的建模，但太多水压计成本过高。本文仅采用4个水压计，能够实现姿态控制）。
\section{Introduction}
%水下机器人很多应用，分析很多应用的原因
Recent years have witnessed the rapid development of underwater robotic systems, thanks to their wide application in aquatic environment protection \cite{application,sensys1},  seabed resources exploration \cite{application0}, ocean mapping \cite{application2}, and so on. 
%Underwater robots are widely applicable because they can cope with the high water pressure and the fluctuating dynamics of ocean currents. These robots are required to perceive complex and perilous marine environments, maintain stability, and accurately position mission objectives, thereby substituting human in such demanding conditions.
These applications highly depend on the localization capability of the underwater robotic platform.
%应用依赖于Localization，给Localization下定义，举例说明
%承上启下

 Localization refers to the process by which the robotic system obtains the position information of a target object. For example, in marine pipeline surveillance \cite{application1},  it is essential for the underwater robot to maintain a stable posture in water and perceive the position of the marine infrastructure. Similarly, the premise of well-performed underwater formation control \cite{application3} is that each robot cost-effectively locates its peers within the collective formation in a distributed way. 

%本文打算设计一个平台 满足三个条件
In this paper, we endeavor to design a cost-effective and practical underwater multi-robot platform equipped with a distributed and robust localization framework. 
To meet the needs of underwater localization in various tasks, a multi-robot platform should obtain the following three capabilities: 
%\begin{itemize}
%多机器人系统中的每个机器人都用昂贵的传感器成本太高了
%\item 

\textbf{1) Cost-effective sensing.} %Underwater robots are required to execute a range of functions, including navigation, obstacle circumvention, route planning, and task fulfillment. These algorithms rely on low-cost perception of the complex underwater environment.
A multi-robot platform is capable of conducting large-scale perceptual tasks concurrently. However, equipping each robot with precise and costly sensors would lead to an excessively high overall cost for the platform. Therefore, each robot should achieve sufficiently effective perception with relatively low-cost sensors.
%集中式的迭代不够鲁棒
%\item 

\textbf{2) Distributed update protocol.} %Each robot in the multi-robot platform should have the capability to locate underwater objects, and subsequently be able to broadcast the positional information of the target object to other robots, enabling a distributed localization of the underwater target object.
In aquatic environments characterized by communication impediments, a centralized protocol can lead to excessive communication pressure on the central node. Moreover, due to the complexity of the underwater environment, the loss of connection with the central node could result in the collapse of the entire platform. A distributed iterative strategy is necessary for ensuring flexibility and robustness of the system.
%定位之前首先要保证自身姿态稳定
%\item 

\textbf{3) Robust orientation control.} 
For a robot to perceive and locate a target object, it is imperative to first ensure the stability of its own orientation. Since the underwater environment replete with currents, turbulence and marine lives, an underwater robotic system needs precise dynamic analysis and the establishment of an appropriate orientation control strategy to counteract potential external disturbances.
%\end{itemize}

% 如果一个机器人平台满足上述三个条件，那么该平台就能够适用于各种各样的场景，实现各行各业的应用。比如，对于某水下物体，多机器人平台可以对其进行分布式追踪。只要目标物体处在平台内任何一个机器人的视野范围内，该机器人就会通过分布式网络，告知其他机器人该物体的位置，实现对该物体的稳定跟踪。再比如，多个机器人在水下场景进行view synthesis任务，各个机器人可以通过相互通信和决策确定每个机器人的最优拍摄位置，最终实现水下场景的高效三维重建。
%If a robotic platform meets the aforementioned three conditions, then it can be applied to a variety of scenarios, achieving various applications that a single robot cannot accomplish. For instance, a multi-robot platform can conduct distributed tracking of a specific underwater object, like an endangered fish or an underwater fighter drone. As long as the target object is within the visual range of any robot within the platform, that robot will communicate the object's location to the others through a distributed network, thereby achieving stable tracking of the object. Furthermore, when multiple robots perform view synthesis tasks in an underwater environment, they can optimize the imaging positions for each robot through mutual communication and planning, ultimately realizing efficient three-dimensional reconstruction of an underwater scene.

% 但是 现有平台不能做到上面三点
However, to the best of our knowledge, no existing underwater robotic platform satisfy all these three requirements. In this paper, we present \textbf{A}n \textbf{U}nderwater \textbf{Ca}mera-Based \textbf{M}ulti-Robot \textbf{P}latform named \textbf{Aucamp}, equipped with low-cost, distributed, and robust localization. Aucamp is the first multi-robot platform systematically designed for distributed localization, by addressing low-cost  perception (monocular-camera-based), designing a distributed framework, and providing robust orientation control mechanisms. 

\subsection{Related Work}
\begin{table*}
  \caption{Existing localization-related sensors that underwater robotic platforms carry}
  \label{tab:sensors}
  \begin{adjustbox}{width=2\columnwidth}
  \begin{tabular}{clllll}
    \toprule
    Sensor type & Representative platforms & Accuracy  & Cost  & Underwater characteristic\\
    \midrule
    Acoustic sensors &\cite{AQUA,sonar1,sonar2,sonar3,sensys2}& High (at long distance) & Very high &  High latency\\
    Electromagnetic wave sensors &\cite{UWB1,UWB2,gps,gnss}& Medium (on land)  & Medium & Invalid underwater\\
    Inertial sensors&\cite{loco,crab}& Low (with cumulative error)  &Low &  Limited to robot localization\\
    Active optical sensors &\cite{lidar,TOF,StructuredLight,sensys3}& High (on land)  & High & Invalid underwater\\
    Binocular cameras &\cite{stereo1,stereo2}& Medium &Low& Insufficiently compact structure\\
   \textbf{Monocular cameras} & \textbf{Aucamp (our work)}& \textbf{Medium} &\textbf{Low}&\textbf{Lack depth information}\\
  \bottomrule
\end{tabular}
\end{adjustbox}
\end{table*}
%现有的各种机器人平台 根据搭载的传感器类型 分类为声学、电磁波、惯性、光学
Underwater robotic platforms are fundamentally related to the development of underwater sensors. According the localization sensors they carry, existing platforms can be categorized into four types: Acoustic sensor-based platforms, electromagnetic wave sensor-based platforms, inertial sensor-based platforms, and optical sensor-based platforms.

%声学传感：高延迟 短距离容易受干扰 昂贵
\textbf{1) Acoustic sensor-based platforms:} Acoustic sensors like sonars are commonly used in multiple underwater robotic platforms \cite{sonar1,sonar2,sonar3,sensys2}. For example,  AQUA \cite{AQUA} use Synthetic Aperture Sonar (SAS) for achieving centimeter-resolution for detailed seabed mapping, and Multibeam Forward-Looking Sonars (FLSs) for obstacle avoidance and navigation enhancement. These sensors can procure the positions of objects and obstacles on the seabed and accomplish localization tasks. 
However, acoustic sensors work by emitting sound waves and measuring the time it takes for the waves to bounce back from objects. Constrained by the speed of sound, their range measurement has a high latency. 
%In addition, acoustic sensors performs better at long distances. For short-range underwater localization, they are prone to interference. 
Besides, high-precision multi-beam acoustic sensors are often prohibitively expensive, usually costing several hundred thousand dollars, which is unacceptable for multi-robot platforms.

%电磁波传感：只能定位自己  在水下不可用
\textbf{2) Electromagnetic wave sensor-based platforms:} Sensors based on electromagnetic waves, such as the Global Positioning System (GPS), Global Navigation Satellite System (GNSS), and Ultra-Wideband (UWB), are specifically designed for localization. Numerous robotic platforms integrate these sensors to achieve self-localization and navigation within their operational environments \cite{UWB1,UWB2,gps,gnss}. 
However, 
%these sensors are only capable of locating the robot itself and cannot be used to perceive unknown objects or obstacles, which greatly limits their application. Moreover, 
due to the absorption of electromagnetic waves by water, sensors based on electromagnetic waves are 
%nearly completely 
attenuated when more than $10\text{cm}$ away from the water surface, which is not suitable for underwater environment.
 
%惯性传感器：只能定位自己  有累计误差
\textbf{3) Inertial sensor-based platforms:} Inertial sensors, i.e., Inertial Measurement Units (IMUs), consist of accelerometers and gyroscopes that measure linear accelerations and angular velocities, respectively. By integrating these measurements over time, the velocity and orientation of an object can be determined. For instance, LoCO \cite{loco} present a low-cost underwater unmanned vehicle that use IMU to obtain the localization of the robot itself. Similarly, bionic cross-media robots like \cite{crab} also utilize IMU to stabilize the orientation of the robot. However, owing to the inherent nature of IMUs to perform integration, platforms that rely on IMUs have significant cumulative errors. Additionally, IMUs can only obtain the positioning of the robot itself and are unable to locate external objects.

%光学传感器：主动传感器会失效  双目的不够紧凑
\textbf{4) Optical sensor-based platforms:} Optical sensors can be further categorized into active optical sensors and passive sensors. Active sensors, which requires the emission of light, are broadly used in existing platforms, including LiDAR \cite{lidar}, 
depth cameras\cite{TOF,StructuredLight},
%Time-of-Flight (TOF) based depth cameras \cite{TOF}, infrared structured light-based depth cameras \cite{StructuredLight}, 
and other custom sensors \cite{sensys3}. These sensors are ineffective underwater due to the absorption effect of water on light. Common passive optical sensors include binocular stereo vision cameras and ordinary cameras. 
Some studies have proposed methods for range and locate objects underwater based on binocular cameras \cite{stereo1,stereo2}. However, due to the need for feature matching, the application of binocular cameras is limited in underwater environments where feature points are sparse. Likewise, the principle of triangulation requires that the two cameras of the binocular system be as far apart as possible, which is contrary to the compactness required for sensors on underwater robots.

The aforementioned analysis of related underwater robotic platforms is summarized in Table~\ref{tab:sensors}. Existing work are inadequate to address the problem of low-cost robust  localization in a distributed way. In this paper, we utilize monocular cameras to design a multi-robot platform and provide cost-effective, distributed, and robust localization. Compared with other robotic platforms, the monocular camera we use is the cheapest and the most compact sensor that can be used for underwater localization. 
%However, there are dual challenges to design such a platform, both from the robot sensors and the external environment.

\subsection{Challenges}
As mentioned in previous subsection, considering the constraints of cost and the underwater environment, we choose monocular cameras for localization. However, a single monocular camera itself cannot achieve underwater localization. To achieve underwater monocular-camera-based localization, it is necessary to address the following four challenges:

%挑战1：单目相机缺少测距能力
\textbf{1) Lack of visual ranging capabilities.} Cameras inherently provide only two-dimensional spatial information.  Monocular cameras are unable to measure distances accurately because they cannot perceive depth from a single image without additional information.
%contextual information.
%or computational techniques. 

%解决方案：
%In this paper, we discern that the variance in clarity across different regions of the image can serve as an indicator of depth information. Based on detailed optical analysis, we establish the intrinsic relationship among the focusing range and the object distance. By employing curve fitting methods, this relationship is made applicable for practical underwater optical ranging. Additionally, we analyze the potential errors of this method and its susceptibility to influences within the underwater environment. Furthermore, based on the imaging model, we derive a localization method based on monocular cameras and conduct experimental validation.

%挑战2：单个相机视野受限
\textbf{2) Limited field of view.} 
For localization purposes, a single monocular camera provides only a relatively limited field of view.
%For a single monocular camera, there is a  trade-off among various imaging parameters. Specifically, an increase in the field of view is often accompanied by a reduction in the shooting distance. The inverse relationship between the field of view and the shooting distance underscores the limited field of view of cameras. 
In the intricate underwater environment, the absence of light sources and the presence of various underwater obstacles can further diminish the actual visual range.
%of the cameras. 

%In this study, a distributed localization iterative architecture is adopted, utilizing multiple robots to form a distributed network that can simultaneously cover a larger area. When one robot detects a target object, it can inform other robots in the platform, and use multiple robots to take multi-angle shots and localize the target object more accurately. 
%Moreover, this global, platform-wide localization approach also facilitates coordinated illumination, potentially employing an artificial multi-source lighting system to address the issue of dim underwater conditions and enhance the effective imaging range of monocular cameras.

%挑战3： 糟糕的水下成像质量
\textbf{3) Poor underwater imaging quality.} The visual quality of images captured underwater is often compromised due to factors such as water turbidity and light absorption. The scattering of light by suspended particles in the water can result in image blurriness and loss of feature. The absorption of light by water at different wavelengths can lead to severe color bias in images. Consequently, %images captured by underwater cameras that are significantly different from those taken above water, thereby rendering 
many image processing algorithms that perform well on land are ineffective in underwater environments. 

\textbf{4) Complex environment.} The underwater fluid dynamics environment is characterized by a high degree of complexity, which poses significant challenges for the control of underwater robots. Because of various currents and vortices, the attitude control of robots is highly vulnerable to interference from the external forces. 

%This paper conduct a detailed dynamic modeling of the designed underwater robotic platform. Based on the dynamics model, we develop a robust orientation control framework. The control system is capable of swiftly restoring the robot to its original horizontal stable orientation after being forcibly altered by external forces. The experimental section of this paper also includes physical validation of the control system.

%Due to the absorption of electromagnetic waves by water, underwater environments often lack adequate lighting and thus rely on artificial illumination. Traditional underwater robotic platforms typically have fixed-mounted lighting and cameras \cite{RelatedPlatform1,RelatedPlatform2,loco} , lacking degrees of freedom (DoF). In this case, the perception system is constrained by the motion system. Consequently, these platforms exhibit insufficient robustness when imaging objects from various perspectives and distances. Increasing the DoF of underwater cameras and illumination systems offers the potential to improve the lighting and imaging capabilities of underwater robotic platforms, which could significantly enhance the efficiency and effectiveness of underwater perception.

In this paper, we undertake many designs to address the aforementioned challenges. For visual ranging, we establish the intrinsic relationship among the focusing range and the object distance, thus achieving depth estimzation of underwater monocular cameras. A distributed update protocol is designed for our multi-robot platform to perceive a broader area simultaneously.% and obtain the object localization more accurately. 
When one robot in our platform detects a target object, it can inform other robots to take multi-angle shots and localize the target object more accurately. 
To improve the imaging quality, we employ a lightweight underwater image enhancement algorithm to enhance the underwater images in real-time. Moreover, to support these designs, a robust orientation control framework is presented to ensure that the robot maintains a stable posture during sensing and localization.

    \vspace{-0.1in}
\subsection{Contributions and Organization}
% 本文工作
In this paper, we present an underwater multi-robot platform for cost-effective distributed localization, named Aucamp. Unlike existing solutions, the underwater multi-robot platform designed in this paper is the first to systematically employ low-cost monocular cameras for underwater localization. Outfitted with a distributed  iteration framework and a resilient control system, the platform is capable of delivering stable and robust localization within a distributed multi-sensor network. This makes it well-suited for a variety of intricate underwater missions, including the tracking of aquatic organisms and the 3D reconstruction of underwater scenes. Our main contributions are summarized as follows:
\begin{itemize}
%我们提出了一套基于相机的水下多机器人平台，提升了水下补光和成像系统的自由度，系统性地实现了基于相机视觉的水下灵活感知和控制。
%针对水下相机测距的难题，创新性地将清晰度相对特征整合进了单目自动对焦相机，并辅以水下光学理论与光线折射矫正算法，低成本地实现了单目水下定位。
%搭配平台设计了分布式的感知算法和鲁棒的控制框架，使得所提出的水下机器人平台能够完成诸如水下生物追踪、水下场景三维重建等复杂任务，并在实验室环境和实际的水体环境进行了初步验证。
\item  We propose an underwater multi-robot platform based on low-cost monocular cameras. To the best of our knowledge, it is the first underwater robotic platform that systematically attains cost-effective, distributed, and robust localization.
\item  To address the challenge of localization with underwater monocular cameras, we integrate clarity features (Tenengrad function) into a monocular camera, supplemented by underwater image enhancement algorithm and underwater camera modeling, achieving low-cost monocular underwater positioning.
\item  A distributed update mechanism and a robust control framework are obtained, enabling the proposed underwater multi-robot platform to accomplish complex tasks such as tracking marine life and 3D reconstruction of underwater scenes. Experimental validations and application instances are also conducted.
\end{itemize}

The rest of this paper is structured as follows: Section 2 provides an overview of our proposed underwater multi-robot platform, including an introduction to the components and the platform workflow. Section 3 focuses on our design on monocular-camera-based underwater perception. Sections 4 and 5 respectively detail the distributed localization protocol and orientation control system of our platform. Section 6 presents experimental evaluation results, coupled with application scenarios like marine life tracking and underwater distributed 3D reconstruction. Finally, Section 7 concludes the paper and discusses potential future research directions.

%####################################################################
%——————————————————overview———————————————
%####################################################################

\section{Platform Components and Implementation}
\begin{figure}[t]
  \centering%\vspace{-0.1in}
\includegraphics[width=0.9\linewidth]{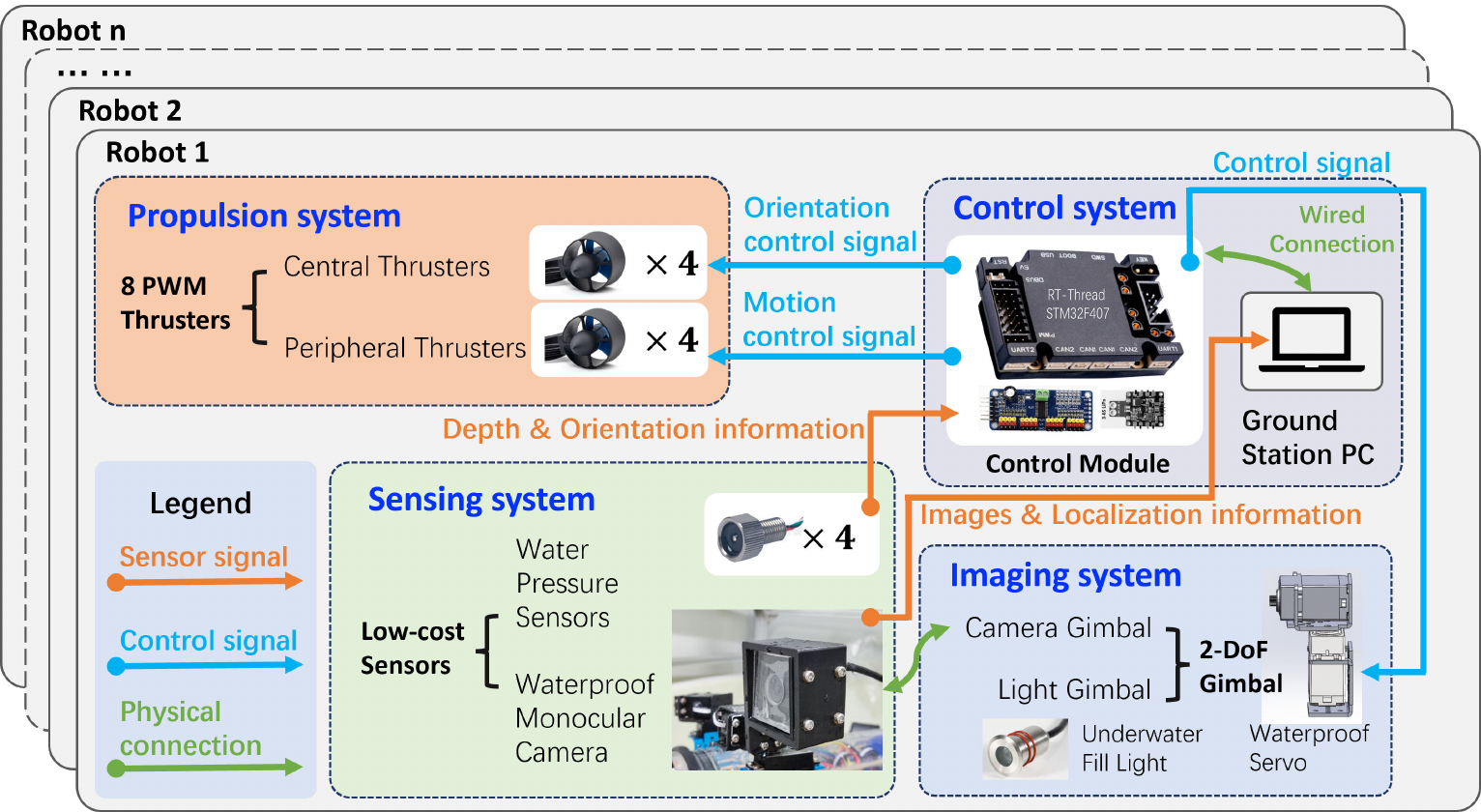}
  
  \caption{Architecture of Aucamp. }
 
\label{Fig:Architecture}\vspace{-0.2in}
\end{figure}

%\textbf{Architecture.}
As illustrated in Figure~\ref{Fig:Architecture}, Aucamp is composed of several identical robots, each equipped with control system, propulsion system, sensing system, and imaging system. We will provide the details of the design and specific implementation of each system, followed by a demonstration of the relationships of these systems using the platform workflow.

%each robot is equipped with eight thrusters, four water pressure sensors, two two-degree-of-freedom gimbals, and a waterproof manual-focus monocular camera. 

%While the platform employs robots of uniform construction, the frameworks proposed in this paper can be easily adapted to new designs involving heterogeneous underwater robot swarms.  

\textbf{1) Control system.}
Each robot in the platform adopts a supervisory and subordinate machine structure, where the supervisory machine is a ground station computer that serves as a medium for communication among robots within the platform; the subordinate machine is the control module of each robot, tasked with aggregating sensor data, executing control algorithms, and dispatching directives to the actuators, ultimately achieving collaborative positioning, planning, and control of the platform. 

To implement such a structure, each robot is endowed with a control module, complemented by corresponding interface expansion, power supply, and information transmission equipment. The control  module is based on the STM32F407 micro-controller, housed within a sealed, waterproof compartment of each robot. All the control modules of every robots are connected to the base station computer via cables. Camera image data, which is too memory-intensive, is processed by the ground computer to derive object positioning data, which is then relayed to the low-level control module. All other sensory data is managed by the control module itself.
The control signals for the actuators are also generated and dispatched by the control module.

% 八推进器
\textbf{2) Propulsion system.}
Propulsion system is designed for the orientation control and the motion control of each robotic body. 
The propulsion system consists of eight thrusters, four of which are fixed in the central part of the robot, and the other four are located at the periphery. The central thrusters are responsible for controlling the orientation and the hovering depth of the robot, while the peripheral thrusters are responsible for controlling the forward, backward, lateral, and rotational movements.

In implementation, all thrusters are regulated by the control system through PWM (Pulse Width Modulation). The thrusters with clockwise rotation and those with counterclockwise rotation are arranged alternately, ensuring that the reaction torque generated by the rotation of the thrusters does not affect the movement of the robot. Each thruster has a distinct initial PWM value. When the PWM input to a thruster exceeds its initial value, it rotates in the forward direction; when it is below the initial value, it rotates in the reverse direction. We control the speed and direction of each thruster by adjusting the delta PWM (the change in PWM value from the initial setting).

% 传感系统
\textbf{3) Sensing system.}
The sensing system of each robot is equipped with a monocular camera to support low-cost distributed localization, and four water pressure sensors for robust orientation control. 
Inspired by \cite{Nature}, pressure sensors can measure the fluid pressure exerted at different locations on the robot, which can be utilized to sense water depth and achieve better posture control. These four pressure sensors are able to obtain orientation data of the robot on two angular axes through pairwise differential measurements. 

 In the specific implementation, each monocular camera is an ordinary camera with a price of less than \$50, capable of manually focusing through control signals from computer (similar to the camera module of a smartphone). Each monocular camera is sealed in an individual transparent acrylic waterproof housing, fixed on a two-degree-of-freedom camera gimbal, and its motion is driven by the movement of the gimbal. The pressure sensing data is transferred to control module to process, and used to control the central thrusters of the propulsion system.

% 二自由度云台、单目相机、补光灯
\textbf{4) Imaging system.}
Each underwater robot in our platform mounts the camera and illumination light on two 2-degree-of-freedom (2-DoF) gimbal and incorporates a waterproof design. When dealing with complex underwater tasks such as tracking rapidly moving underwater objects, this design allows the imaging system to be decoupled from the motion of the underwater robot and controlled independently, thereby providing the illumination and imaging system with greater flexibility.
In the implementation, each gimbal is composed of two waterproof servos arranged vertically and fixed in relation to each other. The servos are control by the control module through PWM.

%\begin{figure}[h]
%  \centering
%  \includegraphics[width=\linewidth]{ZhiyuanLake}
%  \caption{Illustration of xxx. xxx is a multi-robot underwater platform that provide flexible perception and robust control.  (\url{https://www.bilibili.com/video/BV1oC41177Rn}).}
%\label{Fig:ZhiyuanLake}
%%  \Description{A woman and a girl in white dresses sit in an open car.}
%\end{figure}

% 运行逻辑
\textbf{Platform workflow:}
Here is a brief overview of how the various systems coordinate with each other. For orientation control, the sensing system acquires depth information from different positions of the robot and transmits it to the control system, which calculates the orientation of the robot. Based on the orientation information, the control system determines the required rotation speed for the central thrusters and sends it via PWM signals to each central thruster in the propulsion system.

For object positioning, the cameras of the sensing system of every robots obtain positional data of the target object. Subsequently, the control system calculates the necessary angular displacement for the gimbal in imaging system, actuating the gimbal to pivot with PWM and align the camera with the target object. Should the rotation of camera gimbal prove inadequate, the control system will determine the necessary rotational velocity for the peripheral thrusters of the propulsion system, facilitating the movement of the entire robot to align the camera with the target object.

Each robot utilizes its camera-based perception system to ascertain the position of target object. Once a robot within the platform identifies the target object, it estimates the location of the object through its monocular camera localization system and communicates this information with the other robots, initiating distributed location estimation for the object. The proposed distributed update protocol ensures consistency in the location estimation of the object by each robot,  and ultimately achieves distributed localization.

Over the next three sections, we will provide the details of our proposed underwater multi-robot platform, including the approaches to realizing camera-based perception, the cooperative distributed localization algorithm, and the mechanisms for ensuring stable orientation control.

%####################################################################
%—————————————————1 单目定位———————————————
%####################################################################

\section{Monocular-Camera-Based Underwater Perception}
% 在本文设计的平台中，每个机器人的成像系统均仅包含一个低成本的单目相机。从原理上来说，数码相机拿到的本质上是一个RGB三通道的二维向量，该二维向量指示了所成像物体的二维空间信息。但是，在各种应用中，往往需要对物体的三维定位。这意味着估计深度，也即相机到物体之间的垂直距离，成为必需。比如……因此，本章将给出在水下仅用单目相机进行测距的方法
In the platform designed in this paper, each robot is equipped with a cost-effective monocular camera. Essentially, the digital camera provides a two-dimensional matrix across the RGB channels, which indicates the two-dimensional spatial information of the imaged object. 
However, in various applications, three-dimensional localization of the object is often required. This necessitates the estimation of depth, i.e., the orthogonal distance between the camera and the underwater object of interest.
Therefore, this section will present a method for depth estimation using only a monocular camera.

\subsection{Monocular Depth Estimation}

This subsection will specifically elaborate on how to use the regional clarity differences obtained by a monocular camera to derive depth information from images. 

\subsubsection{Underwater image enhancement}
\begin{figure}[h]
  \begin{subfigure}[b]{0.49\linewidth} 
    \includegraphics[width=\linewidth]{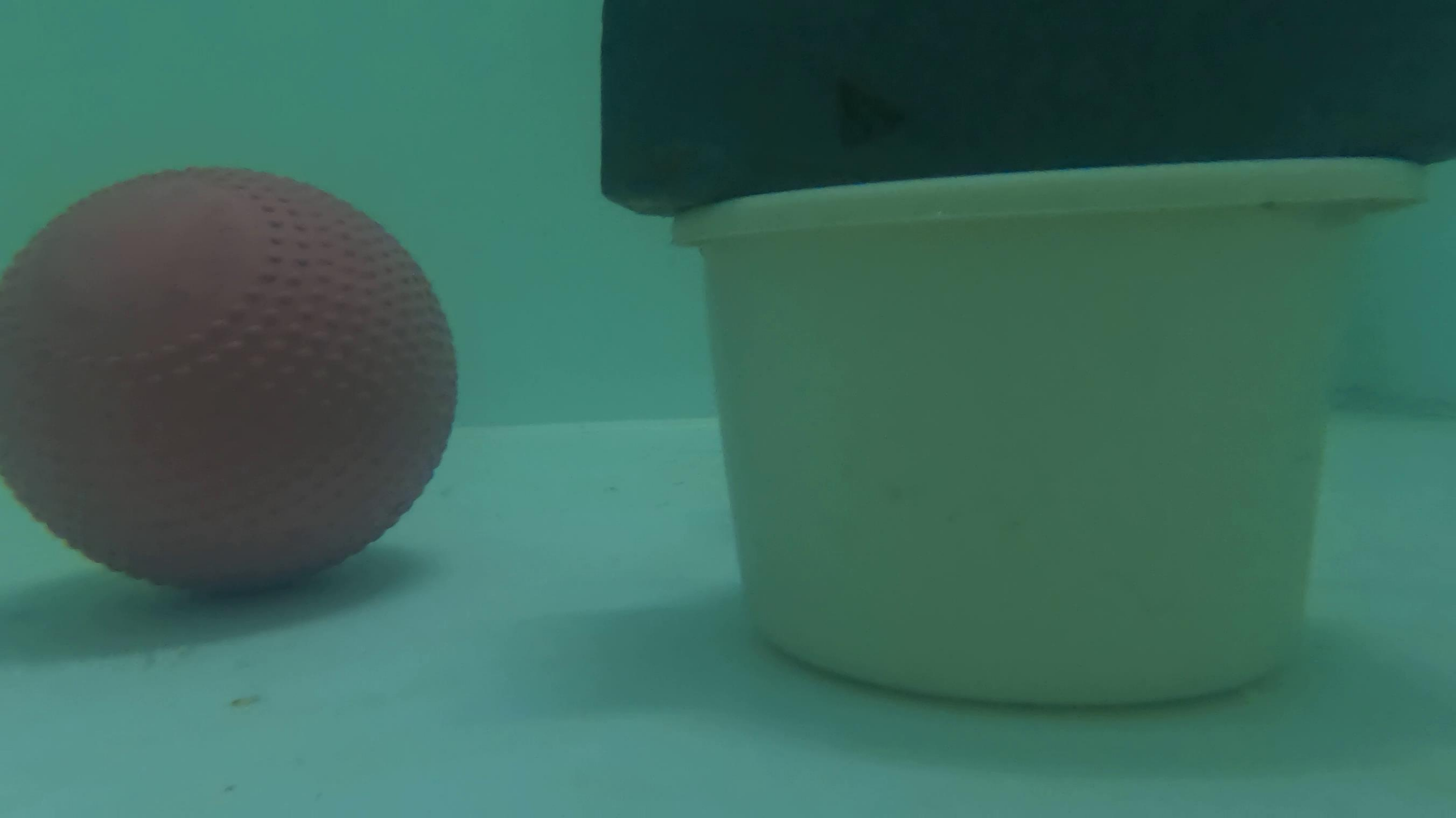}
    \caption{Original image}
    \label{Fig:nerf_origin}
  \end{subfigure}\hfill
  \begin{subfigure}[b]{0.49\linewidth} 
    \includegraphics[width=\linewidth]{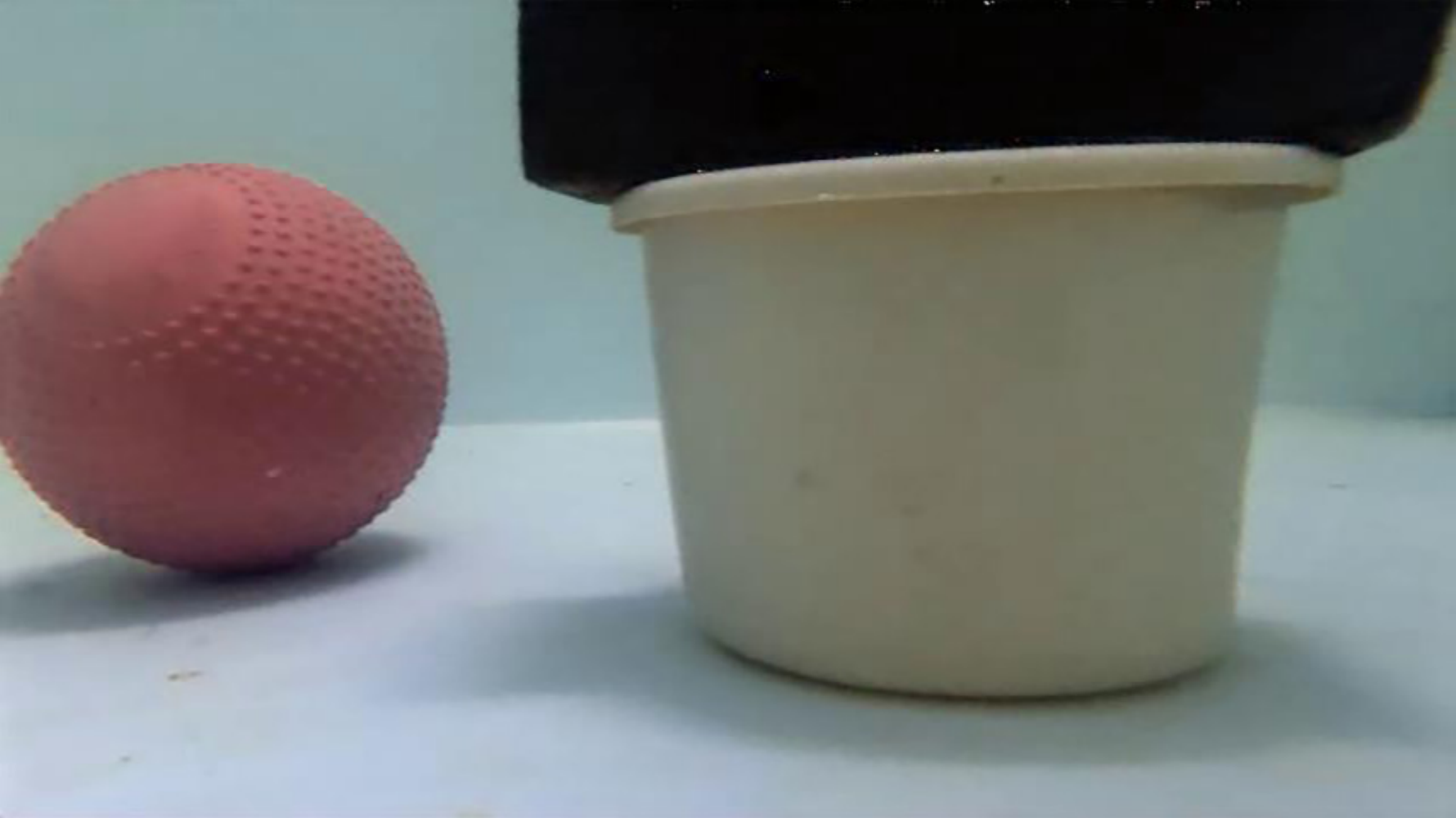} 
    \caption{Enhanced image} 
    \label{Fig:nerf_enhance}
    \label{Fig:now}
  \end{subfigure}
\caption{Underwater image enhancement result.} 
\vspace{-0.2in}
\end{figure}

The original images captured by underwater cameras often exhibit poor quality due to the severe distortion caused by the refraction and absorption of light.
To improve the underwater imaging performance of monocular cameras, we firstly introduce LU2Net \cite{lu2net}, a lightweight underwater image enhancement model, into our platform. Figure~\ref{Fig:nerf_origin} and Figure~\ref{Fig:nerf_enhance} illustrate the original image and enhanced image by LU2Net, respectively.

It is apparent that the image enhancement algorithm we utilize has effectively mitigated the issues of color distortion and blurriness in underwater images, thus supporting subsequent depth estimation procedures. 
Notably, LU2Net operates in real time (up to 100 frames per second), thereby allowing for non-time-consuming integration onto the platform without any negative effect on our perception.

\subsubsection{Clarity feature}
As summarized in Table~\ref{tab:sensors}, most camera-based localization solutions cannot be directly transferred to underwater environments. This is due to their reliance on additional electromagnetic wave (light) emission and reception devices for positioning, which are prone to malfunction underwater.
Existing researches have extensively study the solutions on land. The key reason why they  is that they all rely to some extent on absolute features in the image, rather than relative features. Absolute features are vulnerable to poor lighting condition underwater. To facilitate localization using only a monocular camera, it is essential to extract depth information from the images generated by the camera itself for the purpose of distance measurement.

Our insight reveals that the differences in clarity across various regions in the image represent a relative feature that can be used for distance measurement, and relative features are independent of whether the camera is on land or underwater. This suggests the potential to utilize clarity for underwater monocular depth estimation. %It is noted that in an image captured by a camera, not all areas are in focus. Typically, only objects within a certain distance range appear sharp, while objects that are too close or too far away are blurred. Thus, the clarity within an image serves as a local feature indicative of depth.

\begin{figure}[h]
  \centering
  \begin{subfigure}[b]{0.49\linewidth} 
    \includegraphics[width=\linewidth]{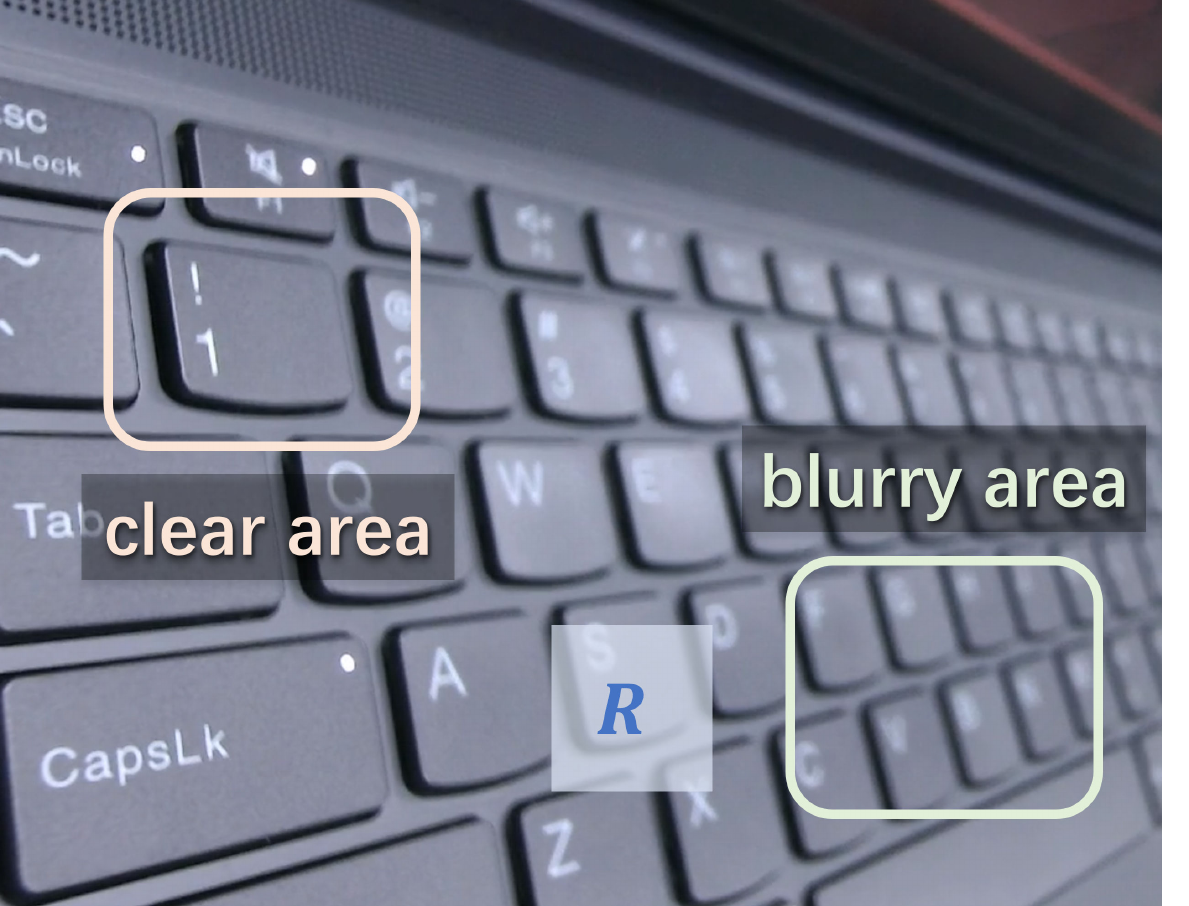} 
    \caption{A schematic diagram to illustrate clarity feature}
    \label{Fig:clarity}
  \end{subfigure}\hfill
  \begin{subfigure}[b]{0.49\linewidth} 
    \includegraphics[width=\linewidth]{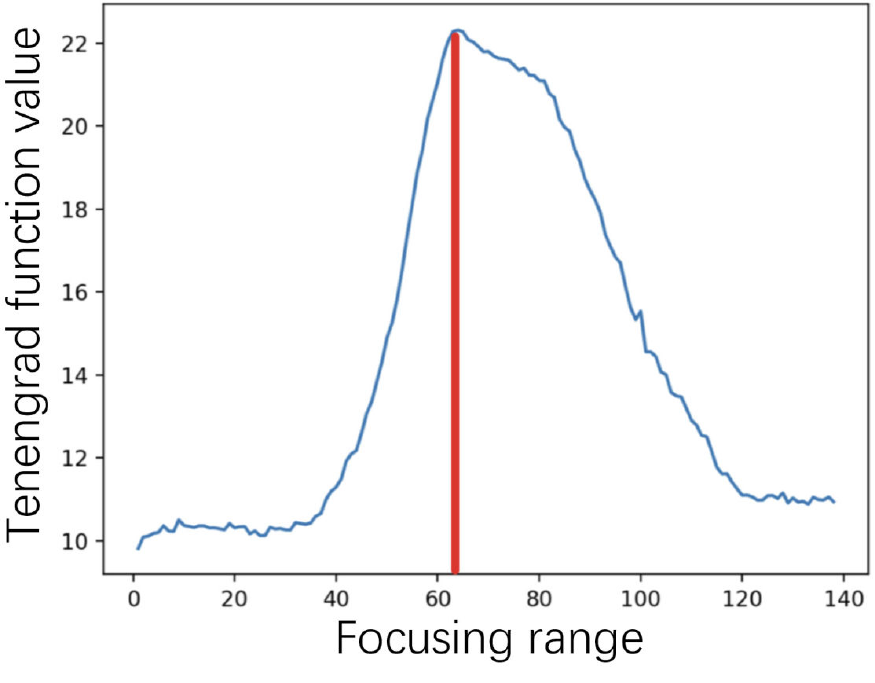} 
    \caption{Tenengrad curve with focus distance variation} 
    \label{Fig:tenengrad}
  \end{subfigure}
\caption{Illustration of Tenengrad clarity feature.} 
\end{figure}
  \vspace{-10pt}

As illustrated in Figure~\ref{Fig:clarity}, there are both clear areas and blurry areas in a image captured by a monocular camera. Apparently, the clear area is closer to the camera than the blurry one, which indicates that clarity feature contains depth information. To apply this idea, we employ the Modified Tenengrad function $T$ \cite{monocular}, an indicator of clarity. For a given region $R$, a higher value of this operator $T(R)$ indicates greater clarity in the imaging of that area. 

For a specific region $R$ within Figure~\ref{Fig:clarity}, one can obtain Figure~\ref{Fig:tenengrad} by changing the focus distance from the nearest to the farthest and calculating the Tenengrad function value for each focus distance. It can be observed that the $T$ is a single-peaked function with respect to focus distance. For a region 
$R$, there exists a unique optimal focus distance. We can utilize this characteristic for monocular distance measurement.

\begin{figure}[t]
  \centering
  \includegraphics[width=\linewidth]{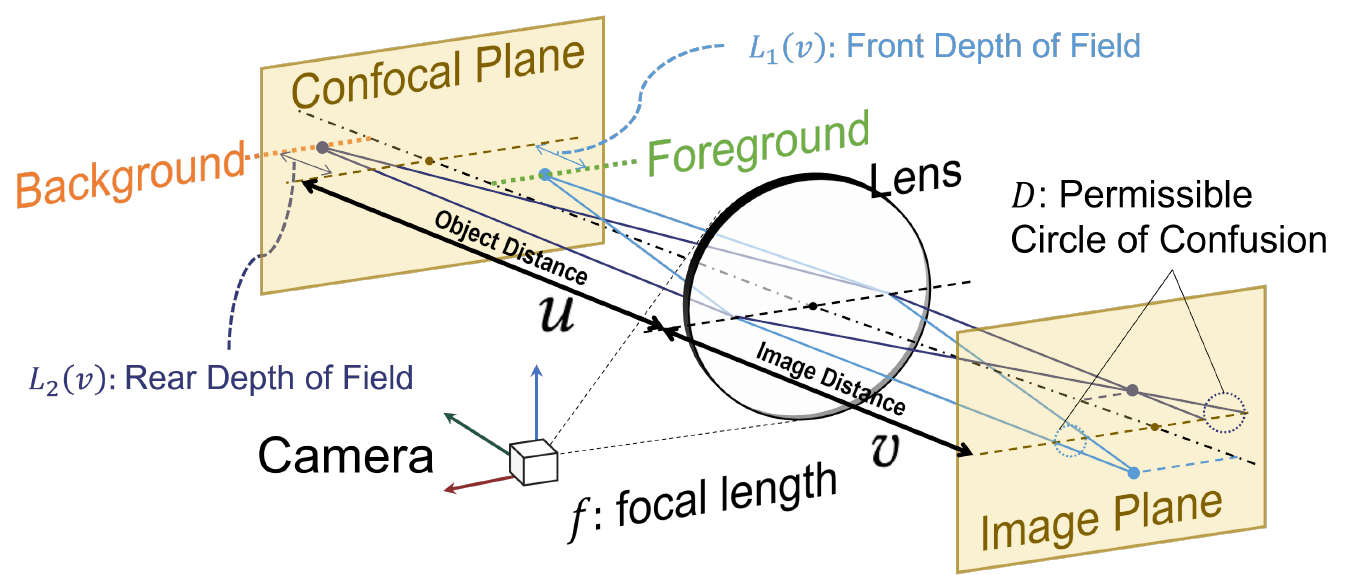}
  \caption{The monocular depth estimation system.}
  \vspace{-0.2in}
\label{Fig:ranging}
%  \Description{A woman and a girl in white dresses sit in an open car.}
\end{figure}

\subsubsection{Monocular distance measurement}

As shown in Figure~\ref{Fig:ranging}, the monocular camera is simplified as a single-lens system. Based on fundamental optics \cite{optics}, we have:
\begin{equation}\label{fvu}
  \frac{1}{f}=\frac{1}{v}+\frac{1}{u},
\end{equation}
where $f$ is the focal length of the camera, $v$ represents the image distance (the distance from the lens to the image plane), and $u$  signifies the object distance (the distance from the object to the lens).
 If $v>f$, \eqref{fvu} is equivalent to
\begin{equation}\label{ufv}
  u=\frac{fv}{v-f}\triangleq u(v),
\end{equation}
which suggests that, given the known focal length $f$ of the camera's equivalent lens, by ascertaining the image distance $v$, one can deduce the object distance $u$, that is, the depth information we required. With a monocular camera, this facilitates the process of three-dimensional object localization.

As mentioned before, $T(R)$ indicates the clarity. When $T(R)$ is at its peak, it implies that the object distance $u_R$ of the objects in that region $R$ conforms to \eqref{ufv}. In this paper, we employ a monocular camera with manual-focus capabilities, which can change the image distance $v$ by adjusting the position $\rho$ of the built-in voice coil motor. Therefore,
\begin{equation*}
  v=g(\rho),\quad\rho^* \triangleq \mathop{\arg\max}_{\rho} T(R),  \quad u_R=\frac{fg(\rho^*)}{g(\rho^*)-f},
\end{equation*}
where $\rho^*$ is the best position of voice coil motor within the camera. Based on our extensive experiments, we claim that the following empirical formula accurately describes the relationship between the variables $u_R$ and $\rho^*$ :
\begin{equation}\label{u_h}
g(\rho)=\kappa\rho, \quad h(\rho^*)\triangleq \frac{f\kappa\rho^*}{\kappa\rho^*-f}+c,\quad  u_R=h(\rho^*)+L(v),
\end{equation}
where $\kappa$ and $c$ are constants determined by the physical characteristic of the monocular camera, $L(v)$ is the error (will be discussed in the next subsection), and the unknown $\kappa$ and $c$ in function $h$ can be fit through experiments. Equation \eqref{u_h} implies that the input-output relationship of the focusing module in the camera to be linear. Furthermore, given the inaccuracy in precisely determining the central position of the equivalent lens of a monocular camera, a constant $c$ is incorporated into \eqref{u_h} to represent the positional offset.

\subsection{Localization with Monocular Camera }
In the previous subsection, we present the approach obtain the object distance $u_R$, i.e., the vertical distance between the monocular camera and object. In this subsection, we use camera imaging model to  obtain the object localization $[x_\text{obj},y_\text{obj},z_\text{obj}]^\text{T}$, and further analyze the theoretical error of this monocular localization.

\subsubsection{Camera imaging model}
\begin{figure}[h]
  \centering
  \vspace{-0.2in}
  \includegraphics[width=\linewidth]{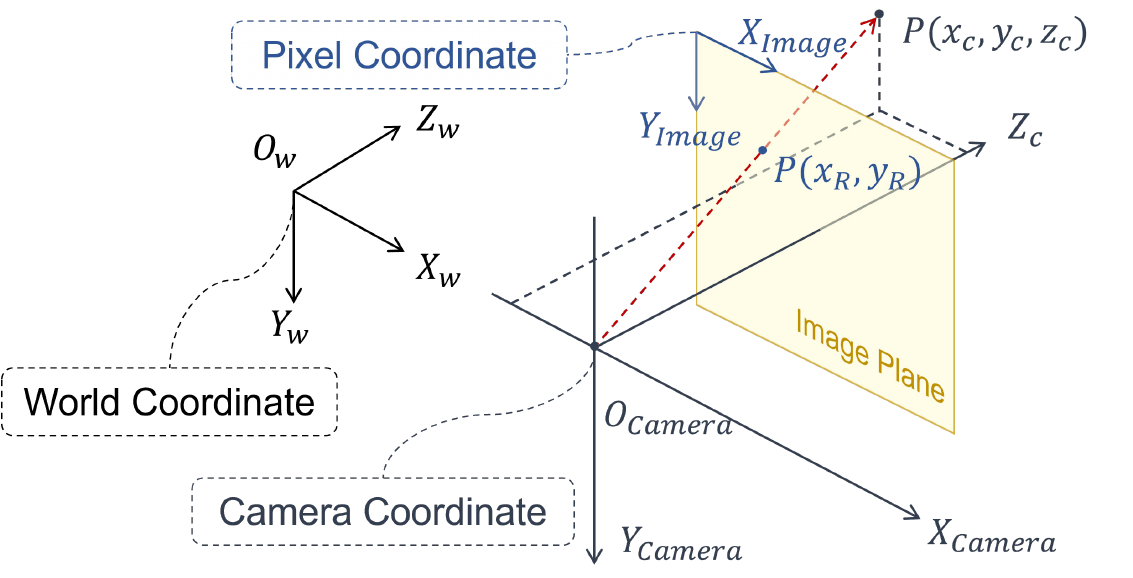}
  \caption{The monocular camera imaging model.}
  \vspace{-0.1in}
\label{Fig:camera}
%  \Description{A woman and a girl in white dresses sit in an open car.}
\end{figure}
As shown in Figure~\ref{Fig:camera}, a calibrated camera can be regarded as a pinhole camera that satisfies the monocular camera imaging model, which can be described by :
$$
[  x_R, y_R, 1]^\text{T}
    = \frac{1}{z_c} \Theta
    [x_c, y_c, z_c]^\text{T},
$$
where \(\Theta\) represents the intrinsic matrix of the camera, $[x_R, y_R]^\text{T}$ is  the pixel coordinates of the object, and $[x_c, y_c, z_c]^\text{T}$ corresponds to the position of the object in camera coordinate system. The intrinsic matrix $\Theta$ is invertible and can be determined using various calibration methods \cite{calibration}. Note that $z_c$ is the vertical distance from the object to the camera, i.e., $z_c=u_R$. Therefore, given the pixel coordinates $[x_R, y_R]^\text{T}$ and the depth $u_R$ the target, the object coordinates in the world coordinate system are:
\begin{equation}\label{xstar}
\hat{x}=
\begin{bmatrix}
        x_\text{obj}\\
        y_\text{obj}\\
        z_\text{obj}
    \end{bmatrix}
    = R_c^O\begin{bmatrix}
        x_c\\
        y_c\\
        z_c
    \end{bmatrix}+p_c^O
=
u_R R_c^O\Theta^{-1}
    \begin{bmatrix}
        x_R\\
        y_R\\
        1
    \end{bmatrix}+p_c^O,
\end{equation}
where $R_c^O$ and $p_c^O$ represent the rotation matrix and the translation matrix from camera coordinate system $O_c$ to the world coordinate system $O_w$. The hat of variable $\hat{x}$ indicates that this value is obtained by the  monocular-camera-based localization presented in this section.
Thus, \eqref{xstar}  provides a method for localization using a single monocular camera.

\subsubsection{Localization error analysis}
The sources of error in the aforementioned localization model are twofold: systematic errors in the depth estimation process, and the impact of the underwater environment on the imaging model.

Our monocular depth estimation approach suggests that the depth of a target object can be ascertained by identifying the focus range in a monocular camera, at the point when the region of that object in the image achieves maximum clarity. 
% 然而，上述方法隐含的假设是一个v对应一个u，也即像距v不变时，仅有像距u为某特定值u(v)的平面上的点在相片中是清晰的。这样的假设过于理想化，未能考虑景深的问题。实际上，由于相机的感光元件（往往被称为CMOS）的最小单元有一定的物理尺寸d（又被称为容许弥散圆），因此在像距为[u(v)-L_1(v),u(v)+L_2(v)]的范围内的所有物体都是清晰的。按照上述方法测量的距离有一定的误差，误差大小即为景深
However, it is based on the implicit assumption that a single image distance $v$ corresponds to a single object distance $u$, meaning that when the image distance $v$ is constant, only the points on a plane at a specific distance $u(v)$ are in focus in the photograph. This assumption is idealized and does not take into account the issue of \textit{depth of field}. 

As illustrated in Figure~\ref{Fig:ranging}, due to the physical dimensions $D$ of the smallest unit of the camera's photosensitive element (often termed  CMOS), all objects within the range $[u(v)-L_1(v),u(v)+L_2(v)]$ appear sharp. The distance measured by our method therefore has a certain degree of error, the magnitude of which is the depth of field $L(v)\triangleq L_1(v)+L_2(v)$. According to \cite{optics}, the measurement error of our method, i.e., the depth of field $L(v)$ satisfy
$L(v)=\frac{f^2u}{f^2-\Gamma D(u-f)}-\frac{f^2u}{f^2+\Gamma D(u-f)}$, where $\Gamma$ is the aperture value of the monocular camera.

Besides, due to the necessity for underwater cameras to be waterproof, an additional waterproof housing is often required compared to cameras used on land.  The housing and the refraction of light by water can affect the monocular-camera-based localization model. According to \cite{Kang}, the impact of the underwater environment on the camera is primarily manifested in the alteration of the intrinsic matrix $\Theta$ and the camera's equivalent focal length $f$.  In our localization approach, the intrinsic matrix $\Theta$ can be corrected by recalibration underwater. The camera's equivalent focal length $f$, which is a coefficient in the ranging function $h$, can also be ascertained through underwater calibration. Therefore, the underwater environment has negligible impact on our localization method.

%####################################################################
%—————————————————2 协同补光———————————————
%####################################################################
\section{Distributed Update Protocol for Global Localization}
% 对于水下机器人集群来说，获得全局定位是困难的。但是，在上一个section中，我们赋予了每个机器人对物体进行单目定位的能力，因此，平台中的每个机器人可以通过和邻居机器人的交互，来使得整个平台对于水下物体有较为精确的的分布式定位。此外，平台也可以根据水下物体的分布式定位，确定每个机器人上搭载的补光灯的朝向，进而实现水下最优补光
% 本章将首先介绍我们平台基本的网络拓扑设定，然后引入Average Consensus算法，最后介绍平台分布式定位和补光的框架流程.
For underwater multi-robot platforms, obtaining global localization is challenging. Not every robot within the platform has access to the target object's field of view in its initial state. Fortunately, as described in the previous section, each robot has been endowed with the capability for monocular object localization. Consequently, provided that at least one robot within the platform has the target object in its field of view, coupled with the distributed iterative strategy presented in this section, the localization information of the target object will propagate throughout the entire multi-robot platform. Therefore, through interaction with neighboring robots, each robot in the platform can contribute to a more precise distributed positioning of underwater objects. % across the entire platform. %Additionally, leveraging the distributed localization data of underwater objects, the platform can determine the orientation of the illumination lights mounted on each robot, thereby achieving optimal underwater illumination.

This section will first introduce the basic network topology settings, then present the update protocol, and finally analyze the global localization process of the platform.

\subsection{Network Topology and Update Protocol}
%在多机器人系统中，多个机器人通过相互通信连接成网络。网络中通常用节点表示机器人，而边表示机器人之间的通信。机器人网络的架构可以分为 centralized、decentralized 、distributed 三种。而distributed
%In multi-robot systems, multiple robots form a network through mutual communication. Within the network, robots are typically represented as nodes, while edges represent the communication between robots. As shown in Figure~\ref{Fig:distributed}, the architecture of robotic networks can be categorized into three types: centralized, decentralized, and distributed.
%
%While centralized systems are easy to manage and control, they lack resilience and scalability. Decentralized systems offer fault tolerance and reduced control, but may lack the unified coordination. Distributed systems, with their ability to scale, balance loads, and provide high availability, are particularly advantageous in large-scale, high-demand environments.
%
%For underwater multi-robot systems, both centralized and decentralized architectures are unsuitable due to their reliance on a central node or key nodes. Underwater communication is challenging, and the central or key nodes bear a substantial communication burden. Moreover, the complex underwater environment is prone to causing damage to the equipment carried by the robots, and the failure of a central or key node often implies the failure of the entire network. In contrast, in a distributed system, the communication load is more evenly apportioned among nodes, and the system is robust against the failure of any single node.
%In this paper, we model our multi-robot platform as a distributed system.

Consider a underwater platform that consists of $N$ robots. The topology of the multi-robot platform can be formulated as an connected undirected graph $G=(V,E)$, where $V=\{1,2,\dots,N\}$ and $E\subset V\times V$ denote the nodes and the edges of $G$, respectively. Each node $i\in V$ represents a robot. Each edge $e_{ij}=(i,j)\in E$ suggest that robot $i$ and robot $j$ can communicate with each other in the platform. Define $N_i\triangleq \{j\in V|(i,j)\in E\}$ as the neighborhood set of robot $i$. The degree of robot $i$ is defined by $d_i\triangleq |N_i|$. 

\begin{figure*}[h]
  \centering
  \includegraphics[width=0.9\textwidth]{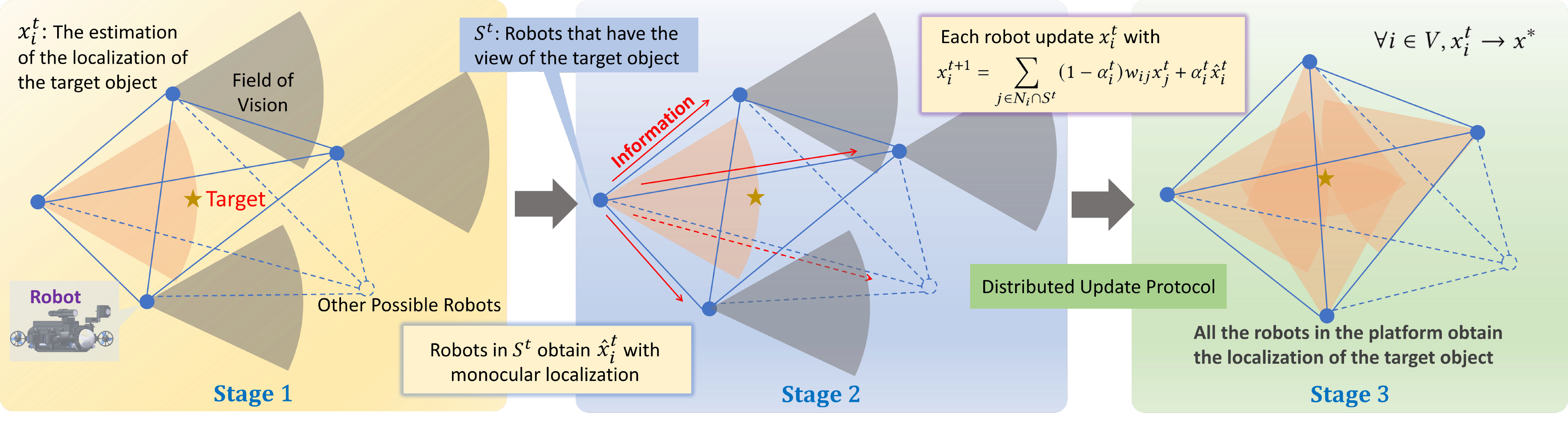}
  \caption{Distributed update protocol.}
  \vspace{-0.1in}
\label{Fig:distributed}
%  \Description{A woman and a girl in white dresses sit in an open car.}
\end{figure*}

In each iteration $t$, based on the method from the previous section, each robot can obtain the position of the target object through monocular depth estimation. This positional information, noted as $\hat{x}_i^t$, will serve as the a prior estimation of the target object's location for each robot. As shown in Figure~\ref{Fig:distributed}, each robot send its $x_i^t$ to all its neighbors, gather information from them, and derive a new a posterior estimation $x_i^{t+1}$ based on its prior estimation $\hat{x}_i^t$ and neighborhood information, thereby achieving a more accurate distributed localization of the target object.

At every iteration $t$, each robot $i$ in the platform maintains its local state variable $x_i^t$, i.e., the estimation for the location of target object. Note that the object of interest may not be within the visual range of every robot, and due to the movement of the robots and the variability of the underwater environment, there is a possibility that the robots may be obstructed or lose track of the target object. Therefore, we introduce the \textit{valid robot set} $S^t\subset V$ for every iteration $t$. All the robots within $S^t$ have the view of the target object. For robots in the set $V\setminus S^t$, since the prior estimation $\hat{x}_i^t$ is meaningless, they determine their local estimation only based on all neighbors in the valid robot set.
% Inspired by Average Consensus algorithm \textbf{[reference]}, the 
 Our update protocol follows
\begin{equation}\label{ac}
 \forall i\in V, t\ge 0, \quad x_i^{t+1}=\sum_{j\in N_i\cap S^t}(1-\alpha_i^t)w_{ij}x_j^t + \alpha_i^t \hat{x}_i^t,
\end{equation}
where $w_{ij}$ the weight, $\forall j\notin N_i, w_{ij}=0$, and the step size $\alpha_i^t$ satisfy  $\forall i\notin S^t,\alpha_i^t=0.$ If $N_i\cap S^t \neq \varnothing$,$\forall i\in V, \alpha^t\le\mathcal{O}(\frac{1}{t}).$ If $N_i\cap S^t = \varnothing, \alpha_i^t=1.$
For $t=0$, we have
\begin{displaymath}
  \forall i \in S^0, x_i^0 := \hat{x}_i^0.
\end{displaymath}

%alpha的引入是借用了e-greedy算法的思想，让每个机器人综合考虑自己基于传感系统的观测和邻居的观测，并且使得alpha逐渐趋近于0，也即逐渐忽略自己的局部观测，进而实现全局一致的分布式定位。
Our protocol is designed based on the \textit{Average Consensus Algorithm} from the field of distributed optimization \cite{consensus} and the \textit{$\epsilon$-greedy Algorithm} from Reinforcement Learning \cite{qlearning}. In \eqref{ac}, the term $w_{ij}x_j^t$ is inspired by Average Consensus Algorithm, which is presented for reaching a common value among a group of agents in a network through local communication and information exchange. The introduction of $\alpha$ borrows the concept from the $\epsilon$-greedy Algorithm, allowing each robot to consider both its own observations based on the sensory system and those of its neighbors. Besides, by making $\alpha_i^t$ gradually approach $0$, the robot $i$ gradually disregard its own local observations, thereby achieving globally consistent distributed localization.
Our distributed update protocol design is summarized in Algorithm~\ref{alg:alg}.

%通过上述的设计，本文提出的迭代式可以让本来没有目标物体视野的机器人通过邻居的信息，计算得到目标物体的大致位置，进而改变相机的朝向，最终获得目标物体的视野。获得目标视野后的机器人就加入了集合S，S逐渐扩大至与V相同，使得所有机器人都获得了目标物体的视野。同时，迭代式能够综合多步多个机器人的观测数据，减小观测的噪声对定位精度的影响。 

\subsection{Global Localization}

Through the aforementioned design, the update protocol \eqref{ac} allows robots that initially lack the view of the target object to calculate  an approximate location through information from their neighbors. Subsequently, they adjust the orientation of their cameras to ultimately obtain a visual field of the target object. Once a robot acquires the target in its field of view, it joins the set $S^t$, which gradually expands until $S^t$ coincides with set $V$, ensuring that all robots have a visual field of the target object. Concurrently, the protocol integrates observation data from multiple steps and multiple robots, reducing the impact of observational noise on positioning.

\begin{algorithm}[!t]
  \SetAlgoLined
  \SetKwInOut{Output}{Output}
  
  \textbf{Initialize} $S^0$\;
	\For{agent $i\in S^0$}{
	$x_i^0 := \hat{x}_i^0$\;
	}
  \For{agent $i\in \mathcal{V}$}{
       \For{$t=0, 1, 2, \cdots, T$}{
	 Try to obtain $\hat{x}_i^t$. Update $S^t,\alpha_i^t$\;
	 \lIf{agent $i\in S^t$}
	 {
	 Send $x_i^t$ to neighbors $j\in \mathcal{N}_i$
	 }
	 Gather $x_j^t$ from all the neighbors $j\in \mathcal{N}_i\cap S^t$\;
	 Calculate $x_i^{t+1}$ with \eqref{ac}\;

  }
  }
  \caption{Distributed Localization}\label{alg:alg}
  
  \Output{$\{x_i^T\}$.}
\end{algorithm}

Noted that the variable $\hat{x}_i^t$ is the measured value of the true position of the target object, we have
\begin{equation}\label{xhat}
\forall i\in V, t\ge 0, \exists \delta, \quad\text{s.t.}   \quad \hat{x}_i^t=x^*+\delta,
\end{equation}
where $x^*$ is the true position of the target object, and $\delta$ represents the measurement noise. Usually, this noise is a zero-mean Gaussian distribution related to the performance of the sensors, i.e.,
\begin{equation}\label{delta}
\delta \sim \mathcal{N}(0, \sigma^2),
\end{equation}
where the variance $\sigma^2$ reflects the error in the measurement. Suppose \eqref{xhat} and \eqref{delta} hold, according to \cite{consensus}, it is easy to prove that when the target object is static in position $x^*$, as $t\rightarrow \infty$, we have
\begin{displaymath}
\forall i\in V,x_i^t\rightarrow x^*.%\triangleq \frac{1}{|S^0|} \sum_{i=\in S^0}x_i^0.
\end{displaymath}

This implies that, as long as the target object is initially within the field of view of any robot within the platform, over time, all robots will orient their cameras towards the object. Moreover, they all will be able to obtain the global localization of the target object.

%####################################################################
%—————————————————3 基于水压计的控制———————————
%####################################################################
\section{Robust Orientation Control}
The preceding sections elaborate the approach to perform distributed localization based on monocular vision. However, the foundation of this method lies in the ability to effectively control the orientation of the robot itself. Otherwise, in the complex underwater environment, if balance cannot be maintained, the camera will not be able to stably capture clear images, the localization algorithms will become ineffective. 

In this section, a comprehensive dynamic analysis of our robotic systems is presented, coupled with the formulation of a well-conceived control system. This ensures that the localization is executed on a platform characterized by a stable and controlled posture.

\subsection{Dynamics Model}
In this subsection, we focus on analyzing how to control the four central thrusters to achieve orientation control of the robot. Given that the robot is capable of stable hovering, the peripheral four thrusters can influence the movements of the robot within the horizontal plane. This aspect, while straightforward, is not discussed  due to space constraints.

\begin{figure}[h]
  \centering
  \includegraphics[width=0.95\linewidth]{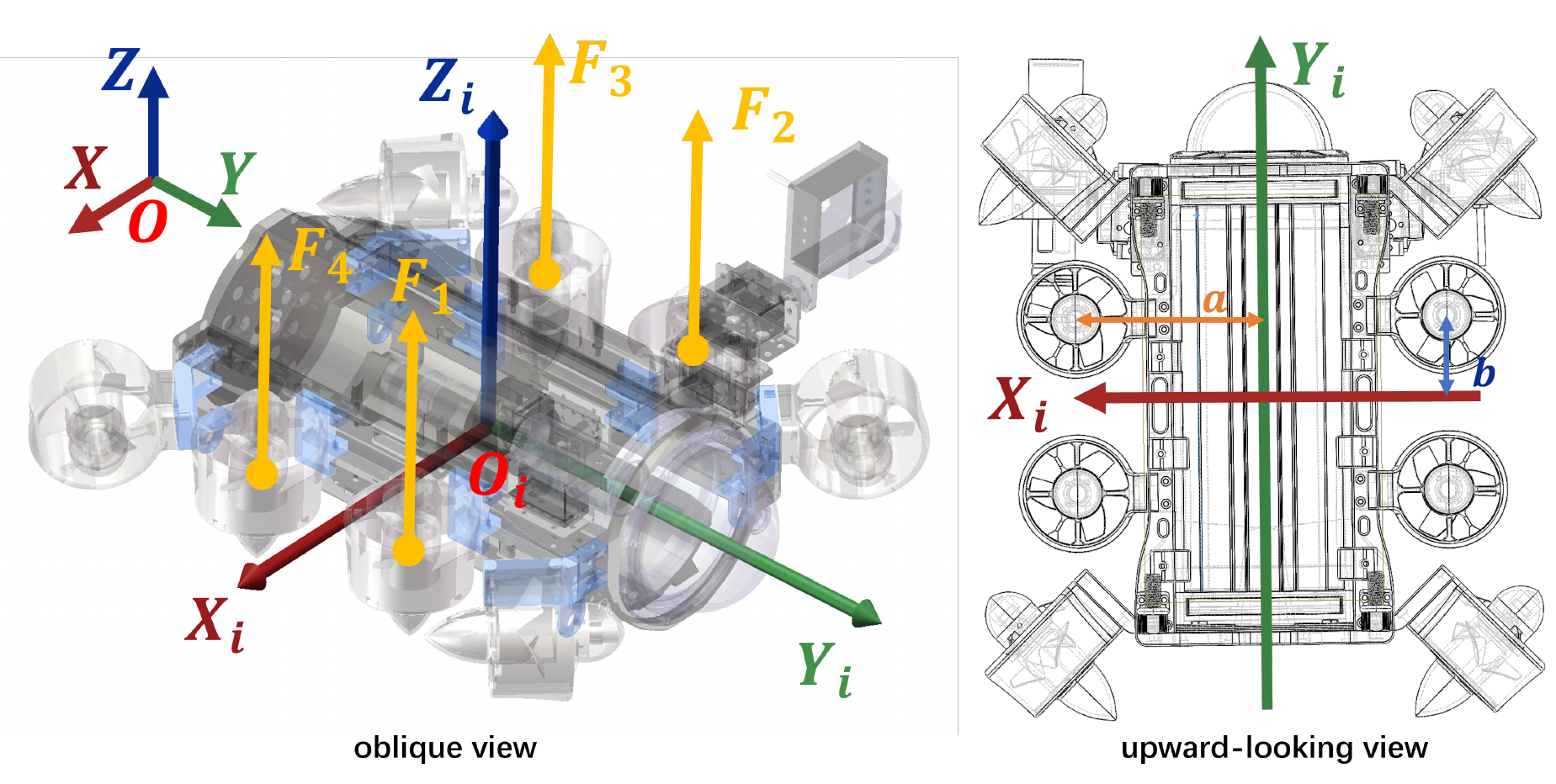}
  \caption{Illustration of the coordinate system.}
  \vspace{-0.1in}
\label{Fig:coordinate}
%  \Description{A woman and a girl in white dresses sit in an open car.}
\end{figure}

% 将世界坐标系设为O(X,Y,Z),每个机器人i的机体坐标系为O_i(X_i,Y_i,Z_i)
As illustrated in Figure~\ref{Fig:coordinate}, the global coordinate system is denoted as $O(X,Y,Z)$, with each robot $i$ having its own body coordinate system  $O_i(X_i,Y_i,Z_i)$. The origin $O_i$   is located at the central position of each robot $i$. Four pressure sensors and the four central thrusters are symmetrically arranged around the origin. The thrusters are positioned at distances $a$ and $b$ from the respective axes. Define the position vector, attitude vector, Euler angle rate vector, of the robot respectivly by 
\begin{displaymath}
\eta=[x,y,z],\xi=[\phi,\theta,\psi],\omega=[\dot\phi,\dot\theta,\dot\psi].
\end{displaymath}
For simplicity, the subscript $i$ indicating different robots are not included. Unless specifically indicated, all dynamic variables in this subsection are set for a single robot $i$. Since the platform employs homogenous robots, the dynamic analysis of one robot is applicable to all. According to Newton's second law of motion, we have:
\begin{equation*}
m\frac{\text{d}^2\eta}{\text{d}t^2}=F_1+F_2+F_3+F_4-[0,0,mg]^\text{T}-F_f,
\end{equation*}
where $\{F_1,F_2,F_3,F_4\}$ represent the thrust from four thrusters, $m$ is the mass of the object, $g$ is the acceleration of gravity, and $F_f$ is the resistance exerted on the robot by the water body. According to fluid mechanism \cite{fluid}, we have
\begin{equation*}
F_f=[k_x\dot x^2,k_y\dot y^2, k_z\dot z^2],
\end{equation*}
where $\{k_x\dot x^2,k_y\dot y^2, k_z\dot z^2\}$ represent the coefficients of fluid resistance in the three axis of $O_i$. Define $\nu_j$ as the rotational speed of the thruster $j$, then $F_j=R_i^O[0,0,K]^\text{T}\nu_j^2(j\in\{1,2,3,4\})$, where $K$ is a coefficient related to the characteristics of the thruster, $R_i^O$ is the rotation matrix from $O_i$ to $O$. 
\begin{align}\left\{\begin{aligned}\label{move}
m\ddot x + k_x\dot x^2&=K\sum_j\nu_j^2(\cos\phi\sin\theta\cos\psi+\sin\phi\sin\psi),\\
m\ddot y + k_y\dot y^2&=K\sum_j\nu_j^2(\cos\phi\sin\theta\sin\psi-\sin\phi\cos\psi),\\
m\ddot z + k_z\dot z^2&=K\sum_j\nu_j^2\cos\phi\cos\theta-mg.
\end{aligned}\right.\end{align}
In the vicinity of the steady state, the torque effect of the fluid on the robot is negligible. Besides, since the actual propellers are installed in an alternating positive and negative rotation pattern, the effect of the blade rotation of the thrusters is also neglected.
By Euler's theorem, the equation of rotation around the center of mass is
$M=I\dot \omega+\omega\times I\omega,$
where 
$$I=
\begin{bmatrix}
I_x & 0 & 0 \\
0 & I_y & 0\\
0& 0& I_z
\end{bmatrix},
M=
\begin{bmatrix}
(\nu_1^2+\nu_2^2-\nu_3^2-\nu_4^2 )b \\
(-\nu_1^2+\nu_2^2+\nu_3^2-\nu_4^2)a\\
0
\end{bmatrix},
$$
thus,
\begin{align}\left\{\begin{aligned} \label{rotate}
I_x\ddot \phi +(I_z-I_y)\dot \theta \dot \psi&=(\nu_1^2+\nu_2^2-\nu_3^2-\nu_4^2 )b,\\
I_y\ddot \theta + (I_x-I_z)\dot \phi \dot \psi&=(-\nu_1^2+\nu_2^2+\nu_3^2-\nu_4^2)a,\\
I_z\ddot \psi + (I_y-I_x)\dot \phi \dot \theta&=0.
\end{aligned}\right.\end{align}
 The equation \eqref{move} and \eqref{rotate} demonstrate the dynamics model of a single robot. In the following section, we will discuss how to utilize this dynamics model for the control system design.

\subsection{Orientation Control}
%在本小节中 我们重点设计控制系统来使得机器人悬浮在水中固定深度，并且保证机器人主平面平行于水平面.由于分布在机器人四周的水压计可以测出当前该位置的深度，可以以水压计的四个输出值z1 z2 z3 z4作为观测量，将中间四个推进器的转速w1 w2 w3 w4作为控制量，建立控制系统
In this subsection, we focus on designing a control system to keep a robot $i$ hovering in water at a fixed depth $z^*$, while ensuring the main plane $O_iX_iY_i$ remains horizontal. Therefore, we are only concerned with the depth and orientation of the robot, i.e., $s\triangleq [z,\phi,\theta,\psi]^\text{T}$. Besides, only the rotational speeds of the four thrusters in the center are controllable, denoted as $\nu\triangleq[\nu_1, \nu_2, \nu_3, \nu_4]^\text{T}$. According to the dynamics model \eqref{move}\eqref{rotate}, it is noted that the variable $s$ is only related to $\sum_j \nu_j^2$, $(\nu_1^2+\nu_2^2-\nu_3^2-\nu_4^2 )$ and $(-\nu_1^2+\nu_2^2+\nu_3^2-\nu_4^2)$. Thus, define control input $U$ with
\begin{equation}\label{u}
\Theta
\triangleq
\begin{bmatrix}
1 & 1 & 1 & 1 \\
1 & 1 & -1 & -1\\
-1& 1 & 1 & -1
\end{bmatrix},U\triangleq\Theta \nu\nu^\text{T}.
\end{equation}

%本文也利用了多个水压计，考虑如何从水压计的数据中得到机器人的姿态信息，进而实现机器人姿态控制。具体而言，首先我们对我们平台的水下机器人的动力学进行建模，分析推进器的动力如何影响机器人的姿态。接着，以多个水压计为传感器，以机器人的推进器转速为控制量，进行控制系统建模，给出基于水压计的鲁棒姿态控制方案。
As for the sensory data, each robot in our platform is equipped with four pressure sensors, each mounted at the four corners of the robot. Pressure sensors are capable of measuring the water pressure at specific locations and obtain the water depth, which is widely used in underwater robots.

The use of multiple pressure sensors can enable more capabilities than simply measuring the depth. Inspired by the multi-pressure-sensor applications in \cite{Nature}, this paper also utilizes multiple pressure sensors to explore how to derive the attitude information from the pressure sensor data, thereby achieving robot orientation control.
Specifically, using multiple pressure sensors as sensors and the speed of the central thrusters as the control quantity, we can model the control system and propose a robust orientation control scheme based on pressure sensors.

The pressure sensors around the robot can measure the depth at their respective positions. Denote the four output values of the pressure sensors as  $\zeta=[\zeta_1, \zeta_2, \zeta_3,  \zeta_4 ]^\text{T}$. Similar to \eqref{u}, define $\mu=\Theta\zeta$. The goal is to make $\mu$ equals to $[4z^*,0,0]^\text{T}$. Define $e\triangleq\mu-[4z^*,0,0]^\text{T}$ and utilize the PI controller, we have $U=K_pe+K_i\int e \text{d}t$. Besides, when the robot is stationary at the target position, gravity is balanced by the thrust force of the four thrusters. Thus, 
\begin{equation}\label{nu}
K\sum_j \nu_j^2=mg.
\end{equation}
Equation \eqref{u} and \eqref{nu} constitute a linear system encompassing four unknown variables,  which can be solved for $[\nu_1^2,\nu_2^2,\nu_3^2,\nu_4^2]$. Ultimately, the necessary rotational speeds of the central thrusters are obtained, thereby achieving robust orientation control of the underwater robot platform.

%####################################################################
%—————————————————实验---------------------———————————
%####################################################################
\section{Evaluation}
In this section, we present extensive experiments to evaluate the performance of various aspects of the underwater multi-robot platform proposed in this paper. %For perception, we will present the curve fitting and depth imaging results of our monocular depth estimation approach. For localization, we will provide a practical example of distributed positioning. Finally, for control, we will present actual pressure sensor data and PWM data of the thrusters to illustrate the robustness of our control framework.

\begin{figure}[h]
  \centering
  \begin{minipage}{0.8\linewidth} 
    \includegraphics[width=\linewidth]{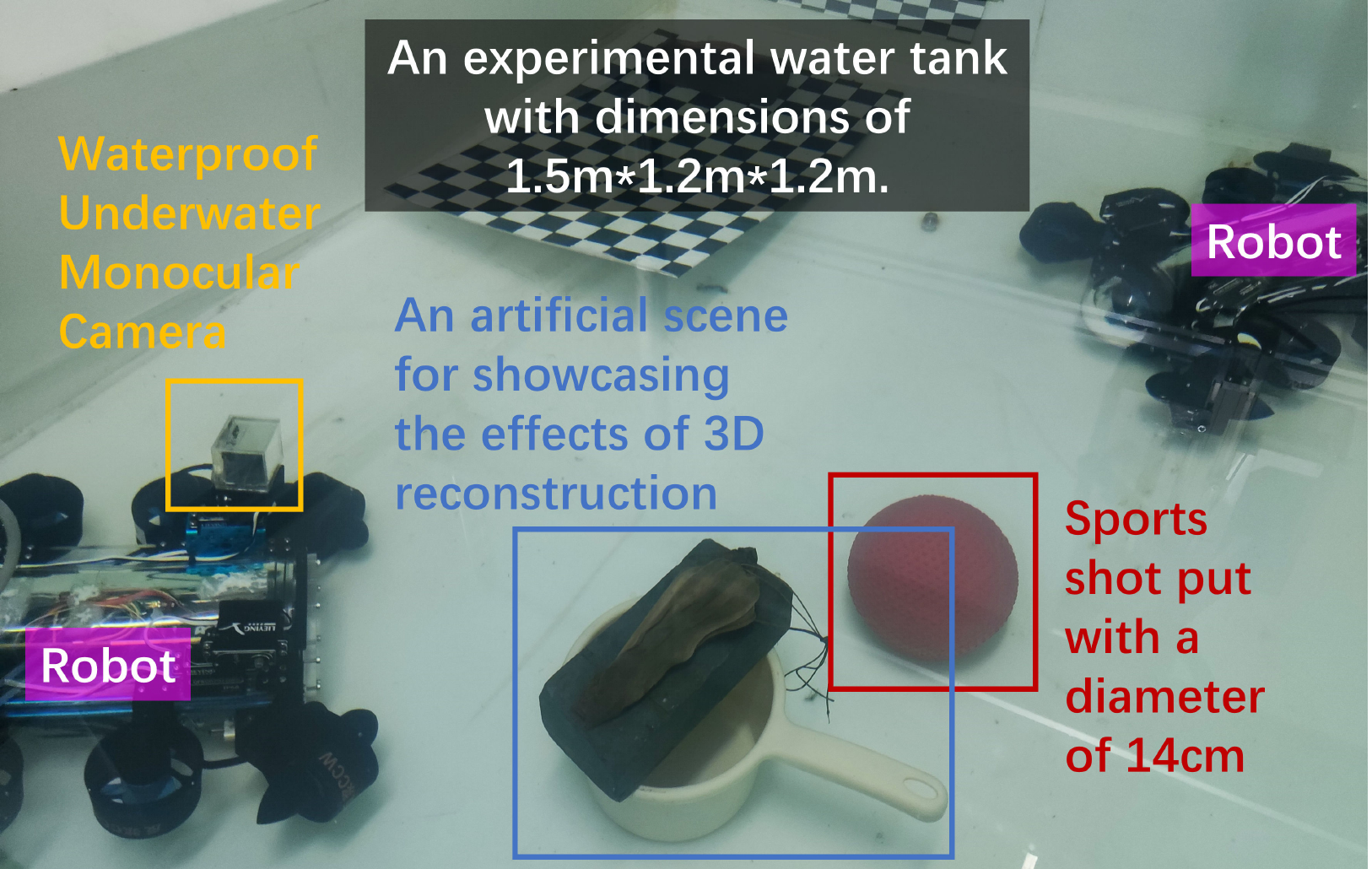} 
    \caption{Test bed settings.} 
    \label{Fig:env}
  \end{minipage}
   \vspace{-0.1in}
\end{figure}

As shown in Figure~\ref{Fig:env}, the experiments of this section are conducted in a water tank measuring 1.5m$\times$1.2m$\times$0.8m, filled with approximately 50cm of tap water. Additionally, there are some auxiliary devices, including a calibration board, a 14cm diameter red sports ball, and a water scoop that is pressed at the bottom of the water by a stone.  Three robots are utilized in the experiments, each connected to a ground computer via cables. The computer serves as the communication medium for the three robots, thereby forming a distributed network.
The distributed network in the various experiments of this section adopts a fully connected topology. 

The specific implementation of each robot can be referred to in Section 2 of this paper.
Various algorithms presented in this paper have certain data that are to be determined, such as the intrinsic matrix of each camera and the transformation matrix from the world coordinate system to the camera coordinate system, all of which are obtained through prior calibration and computation.

%####################################################################
%—————————————————单目深度图--------———————————
%####################################################################
\subsection{Underwater Monocular Depth Imaging}
\begin{figure}[h]
  \centering
  \begin{subfigure}{0.49\linewidth} 
    \includegraphics[width=\linewidth]{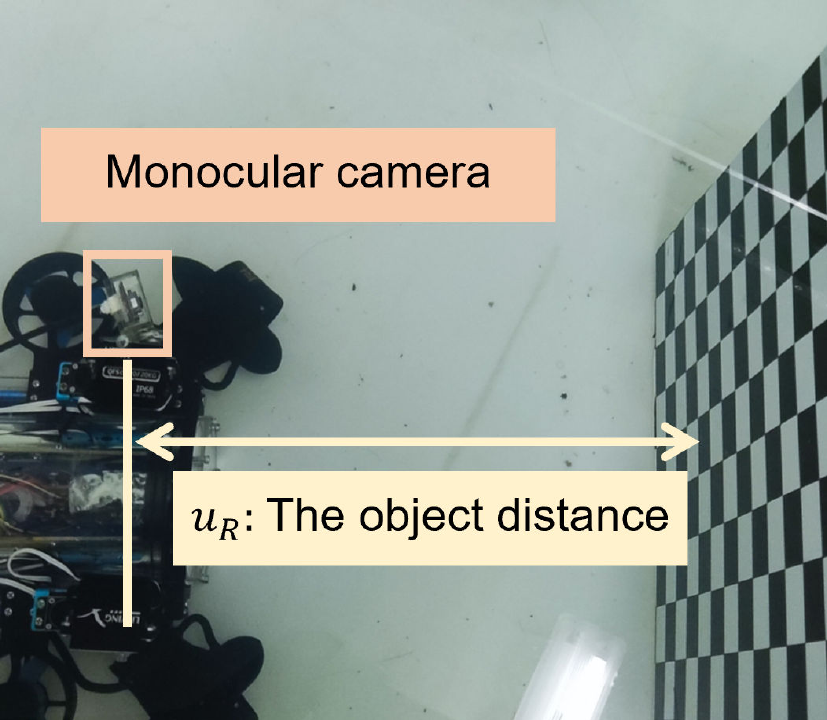} 
    \caption{Experimental setting}   
 \label{Fig:biaoding}
  \end{subfigure}\hfill
  \begin{subfigure}{0.49\linewidth} 
    \includegraphics[width=\linewidth]{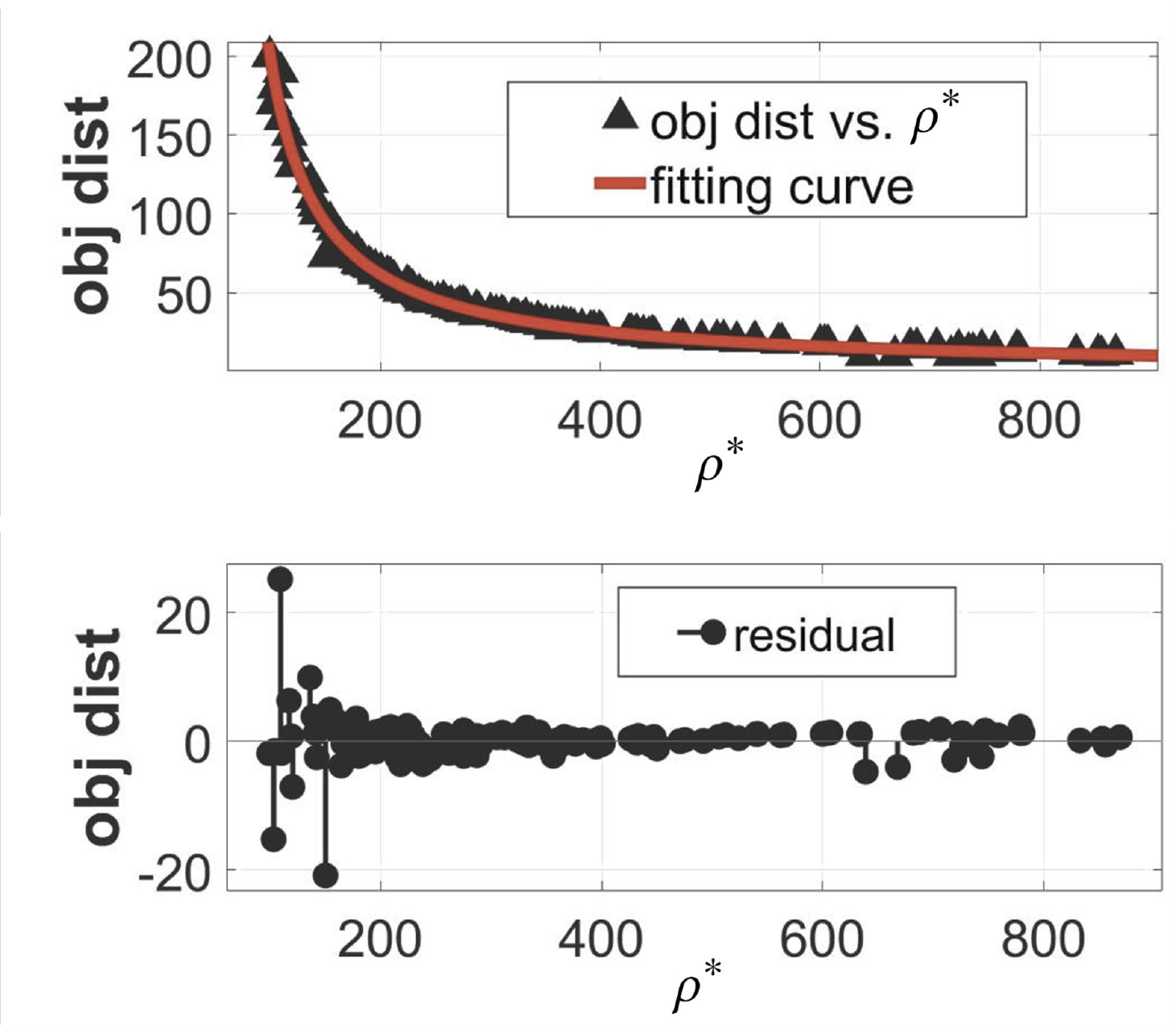}
    \caption{Curve fitting result} 
    \label{Fig:cftool}
  \end{subfigure}
\caption{Clarity function calibration experiment.}
\vspace{-0.1in}
\label{Fig:cali}
\end{figure}

To implement our underwater monocular depth imaging method in practical scenario, the first step is to obtain the function $h$ in \eqref{u_h}. The Figure~\ref{Fig:cali} demonstrates the experimental setup and results of this model calibration process. 

As shown in Figure~\ref{Fig:biaoding}, by altering $u_R$ the distance between the calibration board and the camera, a series of optimal focal values $\rho^*$ and corresponding object distances $u_R$ are obtained. These data are then put into curve fitting toolbox to determine the approximation of $h$. According to the fitting results (shown in Figure~\ref{Fig:cftool}), we have $h=\frac{kxf}{kx-f}+c$, where $k=0.3922,f=0.7431,c=0.7577$. The R-squared of the regression is $0.99$, with $RMSE=3.4285$.

The fitting results support our empirical formula proposed in \eqref{u_h}, thereby validating the feasibility of our approach to achieve monocular distance measurement based on clarity. 
%Each camera undergoes the this regression analysis and calibration procedure, with the resulting expressions being utilized for the purpose of monocular localization.

\begin{figure}[h]
  \centering\vspace{-0.1in}
  \begin{subfigure}{0.49\linewidth}
    \includegraphics[width=\linewidth]{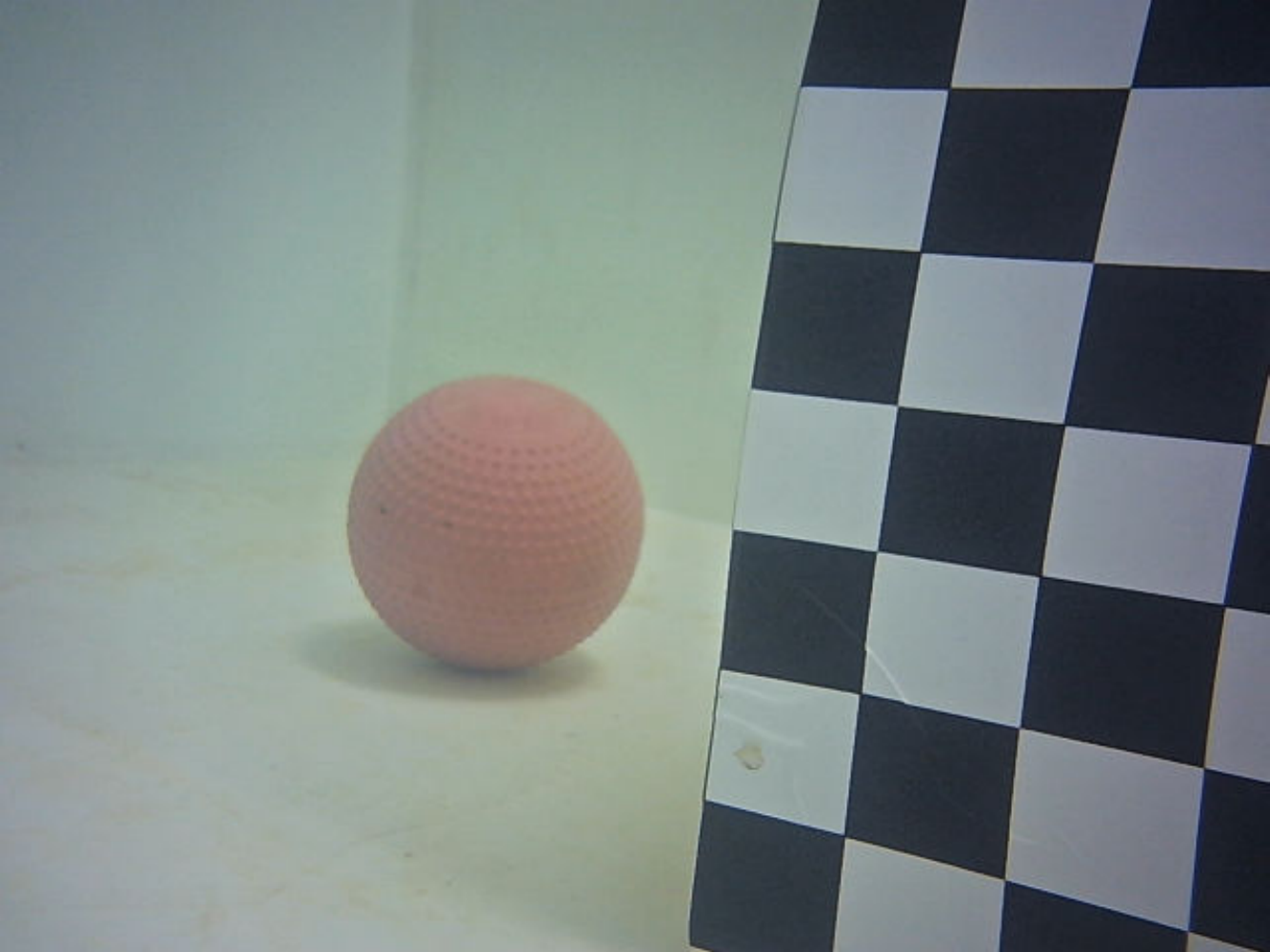} 
    \caption{Enhanced Image}
    \label{Fig:now}
  \end{subfigure}\hfill
  \begin{subfigure}{0.49\linewidth} 
    \includegraphics[width=\linewidth]{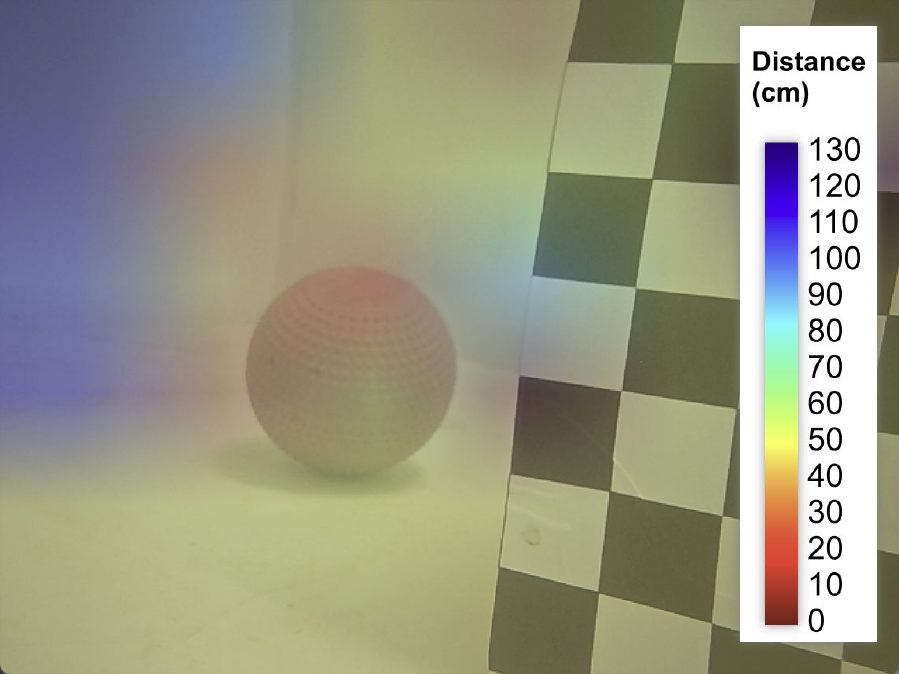} 
    \caption{Monocular depth imaging}
    \label{Fig:colored}
  \end{subfigure}
\caption{Monocular depth estimation result.}\vspace{-0.1in}\label{Fig:dep_res}
\end{figure}

Figure~\ref{Fig:dep_res} illustrate the depth estimation result using the aforementioned calibrated function. In experiments, the images captured by the camera are enhanced (Figure~\ref{Fig:now}) in real-time through the LU2Net algorithm \cite{lu2net}.  To obtain depth information across the entire field of view, the camera images are partitioned into blocks of 50x50 pixels. The clarity function Tenengrad is computed in each block to determine the optimal focus value in each block. The depths are then calculated and visualized (Figure~\ref{Fig:colored}). It can be observed that this method effectively handles flat surfaces lacking distinctive features, 
such as the edges and bottom of the tank. Our method distinguishes between them and successfully measures their exact depth.

%However, due to the small size of the partitioned areas, the results of the Tenengrad function within a 50x50 pixel range may not be accurate, resulting in some erroneous areas in the image. To enhance the ranging performance, a rational segmentation algorithm needs to be designed, which will also be one of our future  work.

%####################################################################
%—————————————————分布式定位实验-------——————————
%####################################################################
\vspace{-0.1in}
\subsection{Distributed Underwater Localization}
In this subsection, we choose a sports shot put (lying on the bottom of our experimental tank) as the target object to implement our distributed localization protocol. In the experiment, each robot utilizes the orientation control algorithm proposed in Section 5 of this paper to hover at a depth of approximately 40 cm in the water, using the camera gimbal mounted on the robot to locate the target object at the corresponding position. 

\begin{figure}[b]
  \centering
  \vspace{-0.1in}
  \begin{subfigure}{0.49\linewidth} 
    \includegraphics[width=\linewidth]{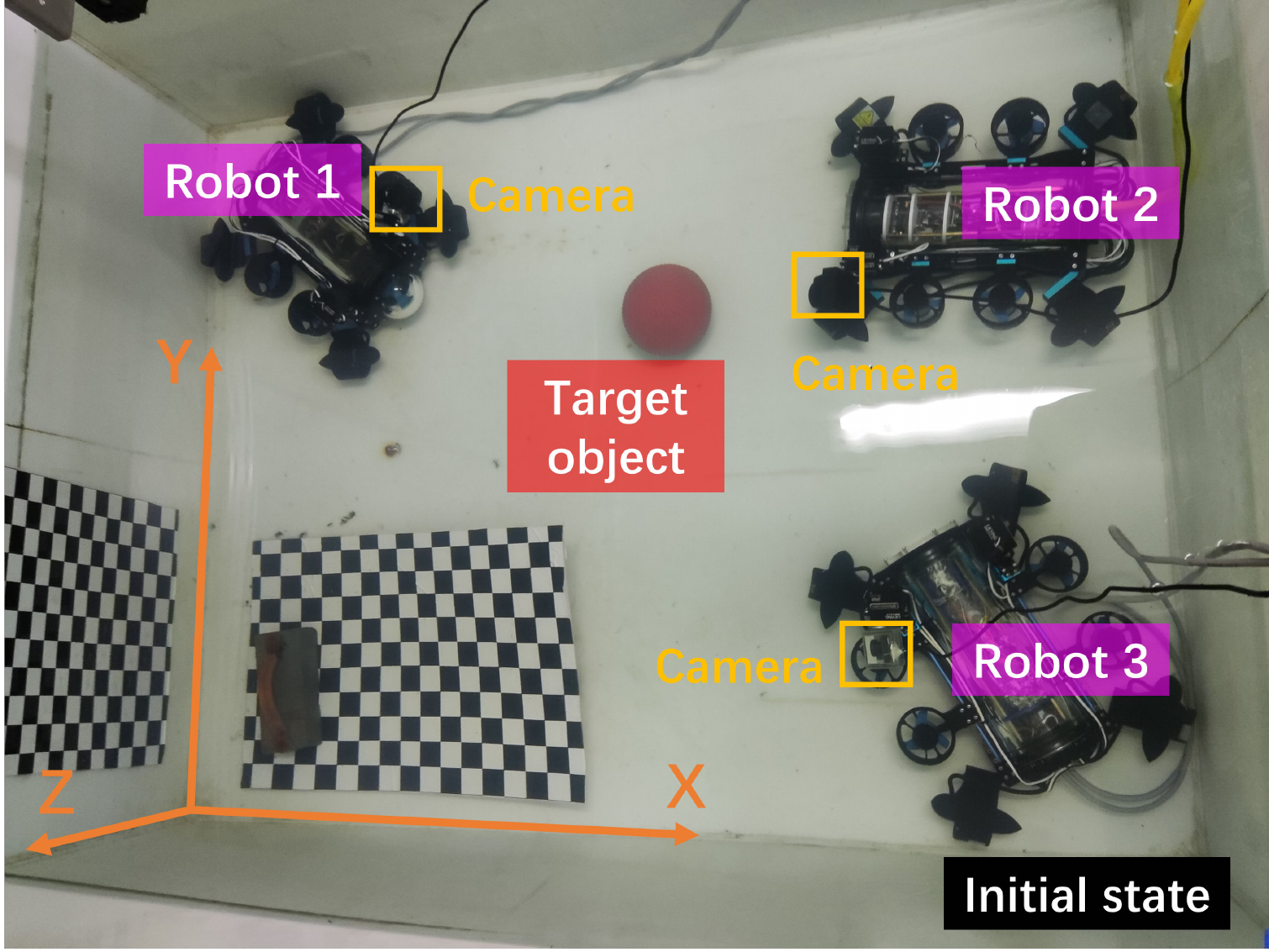} 
    \caption{Initial state} 
    \label{Fig:initial}
  \end{subfigure}\hfill
  \begin{subfigure}{0.49\linewidth}
    \includegraphics[width=\linewidth]{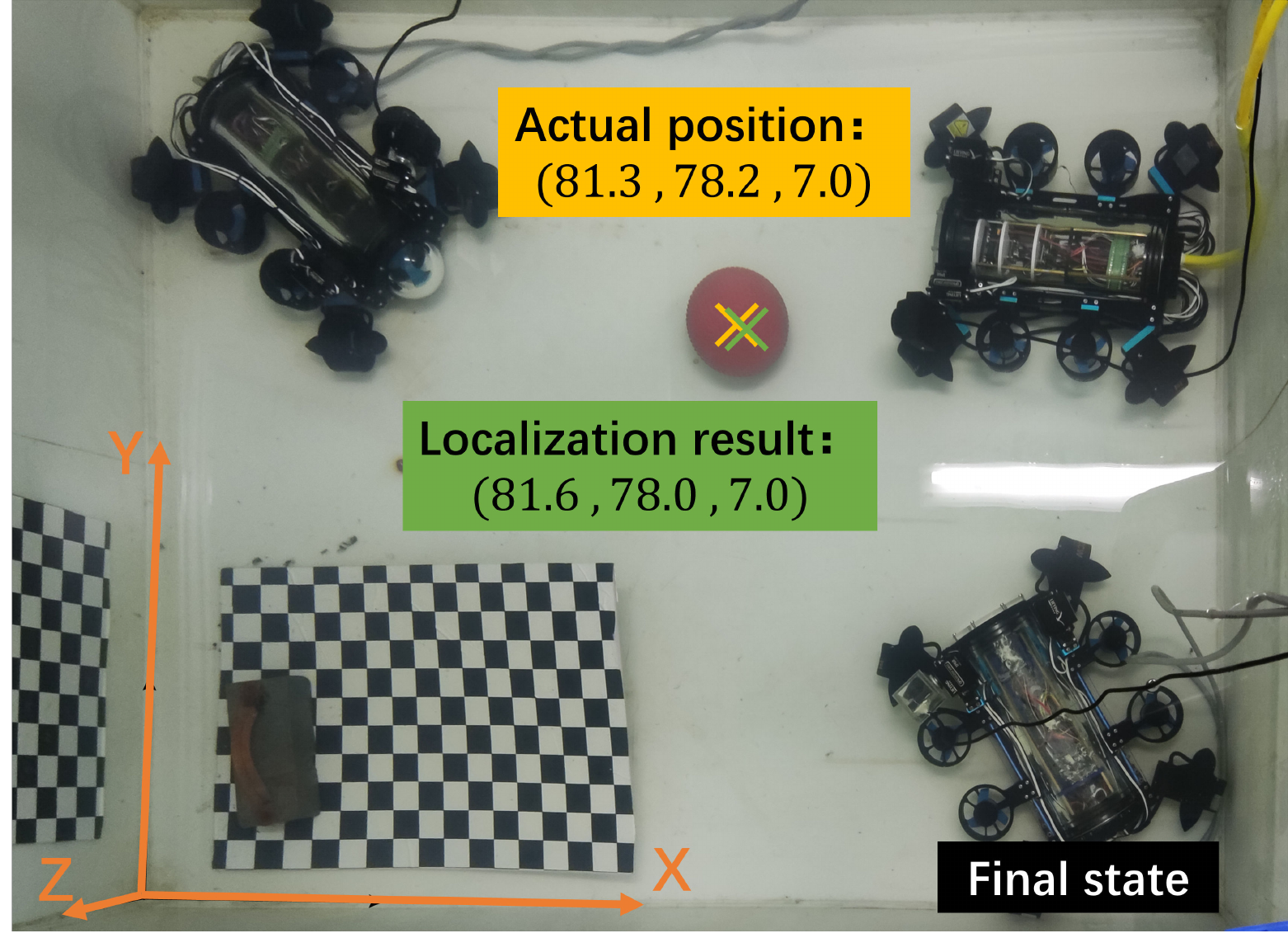}
    \caption{Final state} 
    \label{Fig:final}
  \end{subfigure}
  \begin{subfigure}{0.49\linewidth} 
    \includegraphics[width=\linewidth]{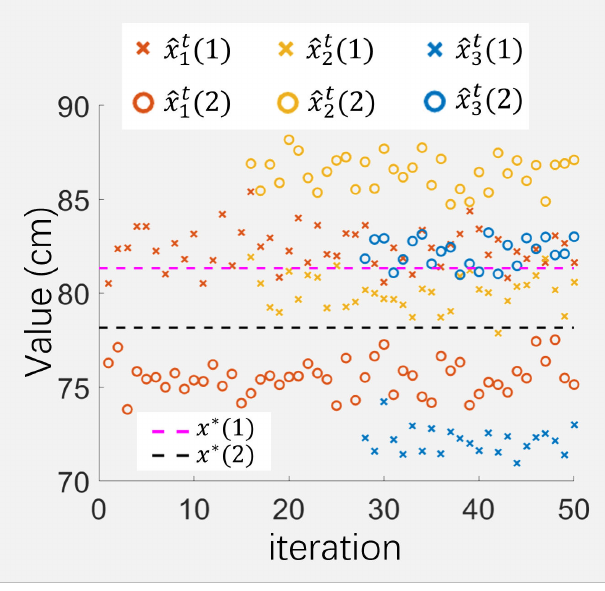} 
    \caption{The first two dimensions of $\hat{x}_i^t$ (Non-distributed protocol)} 
    \label{Fig:3}
  \end{subfigure}\hfill
  \begin{subfigure}{0.49\linewidth}
    \includegraphics[width=\linewidth]{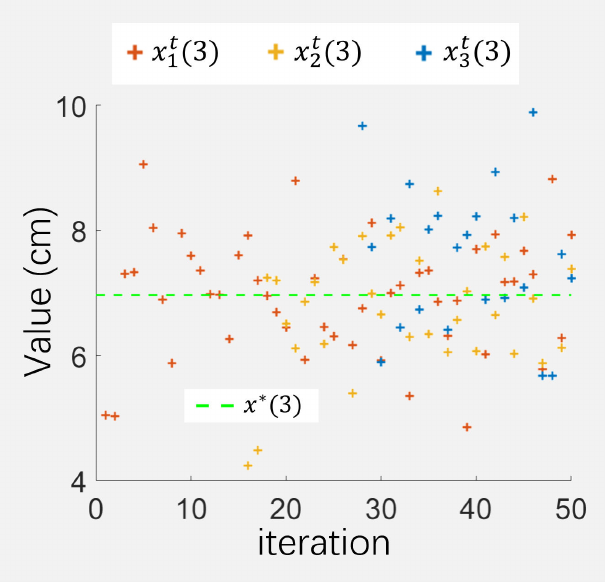}
    \caption{The third dimension of $\hat{x}_i^t$ (Non-distributed protocol)} 
    \label{Fig:4}
  \end{subfigure}
  \begin{subfigure}{0.49\linewidth} 
    \includegraphics[width=\linewidth]{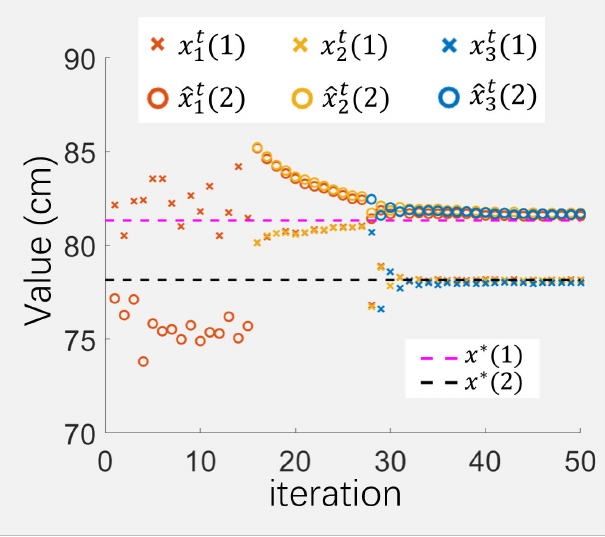} 
    \caption{The first two dimensions of ${x}_i^t$ (Distributed localization)} 
    \label{Fig:1}
  \end{subfigure}\hfill
  \begin{subfigure}{0.49\linewidth}
    \includegraphics[width=\linewidth]{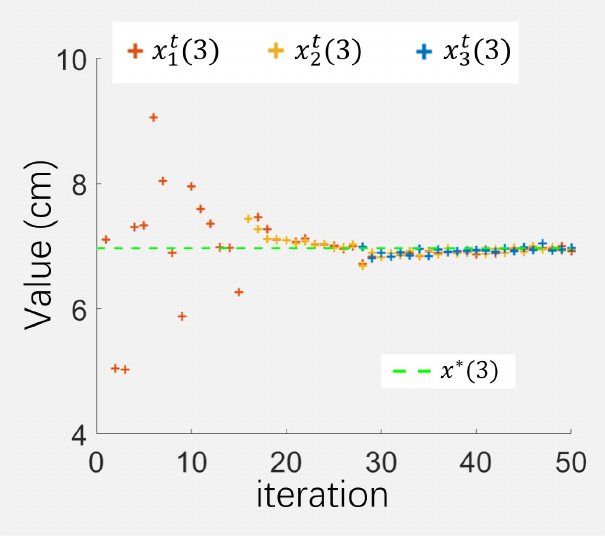}
    \caption{The third dimensions of ${x}_i^t$ (Distributed localization)} 
    \label{Fig:2}
  \end{subfigure}
  \caption{The distributed localization results.}
  \vspace{-0.2in}
  \label{Fig:distributed_res}
\end{figure}

As shown in Figure~\ref{Fig:initial}, the world coordinate system is anchored at the lower left corner of the tank's base. In the initial state, robot 1 has the target object within its field of view of camera and is able to obtain the localization estimation of the target object, while robots 2 and 3 do not acquire the position of the object.

The distributed localization results are visualized in Figure~\ref{Fig:distributed_res}. In the figure, the numbers within parentheses denote the dimensions. For example, $\hat{x}_2^t(1)$   represents the first dimension of $\hat{x}_2^t$, i.e., the prior estimation of robot $2$ at iteration $t$. Due to the significant disparity between the first two dimension ($XY$-coordinates)  and the thired dimension ($Z$-coordinates), the visualization is divided into two sets. % for better visual effect. 

It can be observed that at first, only robot 1 has data $\hat{x}_1^t$, i.e., the estimated localization of the target object. The positional information of the target object is then continuously conveyed to the other two robots through the iterations of our distributed network. Based on the localization information, robots 2 and 3 adjust their camera gimbals to search for the target object. They obtain the view of the target object in the 16th and 28th iterations, respectively. After acquiring the view, the robots joined the valid robot set $S^t$. Consequently, the distributed protocol we designed also incorporated their measurement information into the network. Ultimately, we achieved a relatively accurate distributed localization with an error margin only at the millimeter level.

Figure~\ref{Fig:3} and Figure~\ref{Fig:4} are the prior estimation of the target object, i.e., the localization results without distributed protocol. Figure~\ref{Fig:1} and Figure~\ref{Fig:2} are the distributed localization results, which are apparently better than that without distributed protocol, which strongly demonstrate the effectiveness of the algorithm proposed in Section 4.

Figure~\ref{Fig:final} shows the final state of the experiments. The relative error is \textbf{less than 0.4\%}. These experiments validate the effectiveness of the distributed localization mechanism of our platform.

%####################################################################
%—————————————————鲁棒控制实验-------——————————
%####################################################################
    \vspace{-0.1in}
\subsection{Robust Orientation Control}

\begin{figure}[h]
  \centering
  \begin{subfigure}[b]{0.49\linewidth} 
    \includegraphics[width=\linewidth]{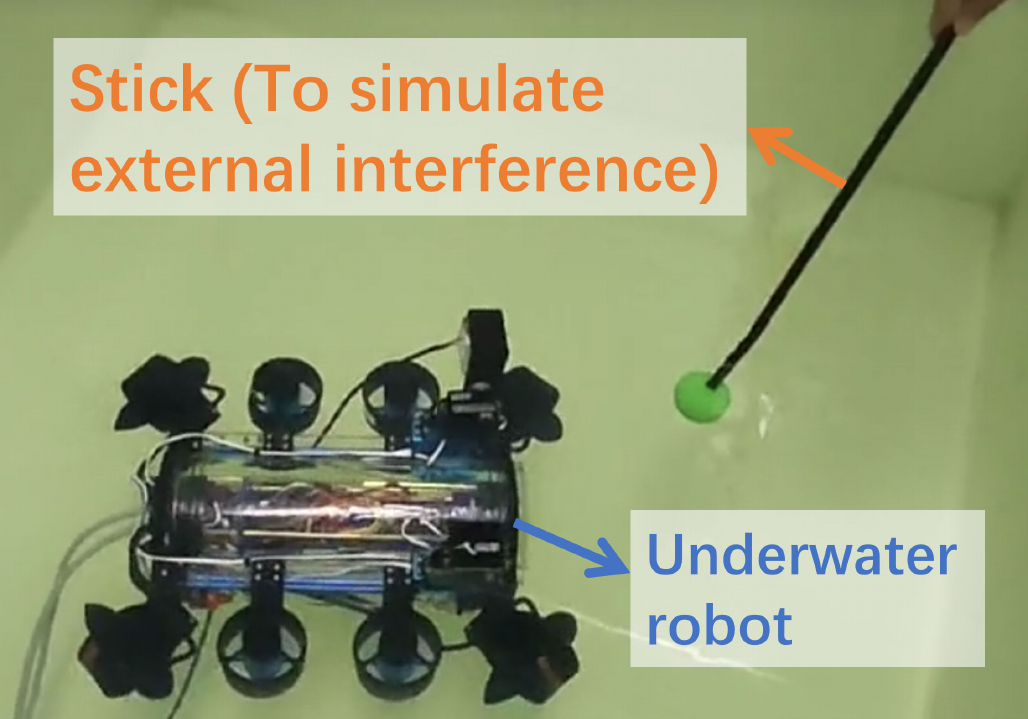}
    \caption{Experimental scenario }
    \label{Fig:stick_env}
  \end{subfigure}\hfill
  \begin{subfigure}[b]{0.49\linewidth} 
    \includegraphics[width=\linewidth]{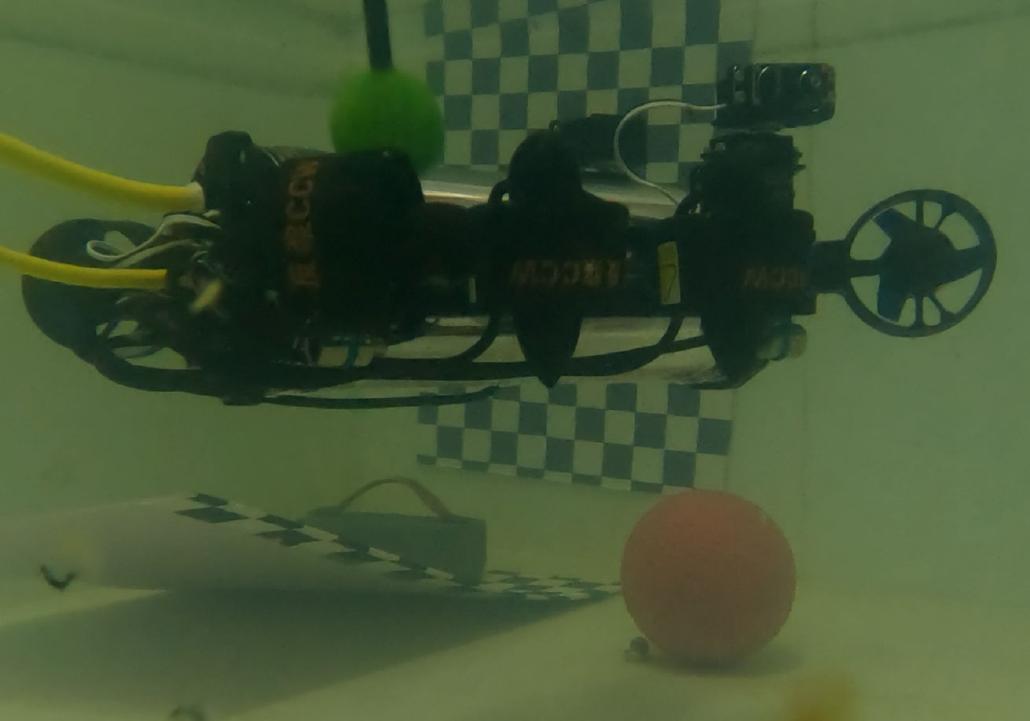}
    \caption{Push the robot with a stick}
    \label{Fig:stick}
  \end{subfigure}  \hfill
  \begin{subfigure}[b]{0.49\linewidth} 
    \includegraphics[width=\linewidth]{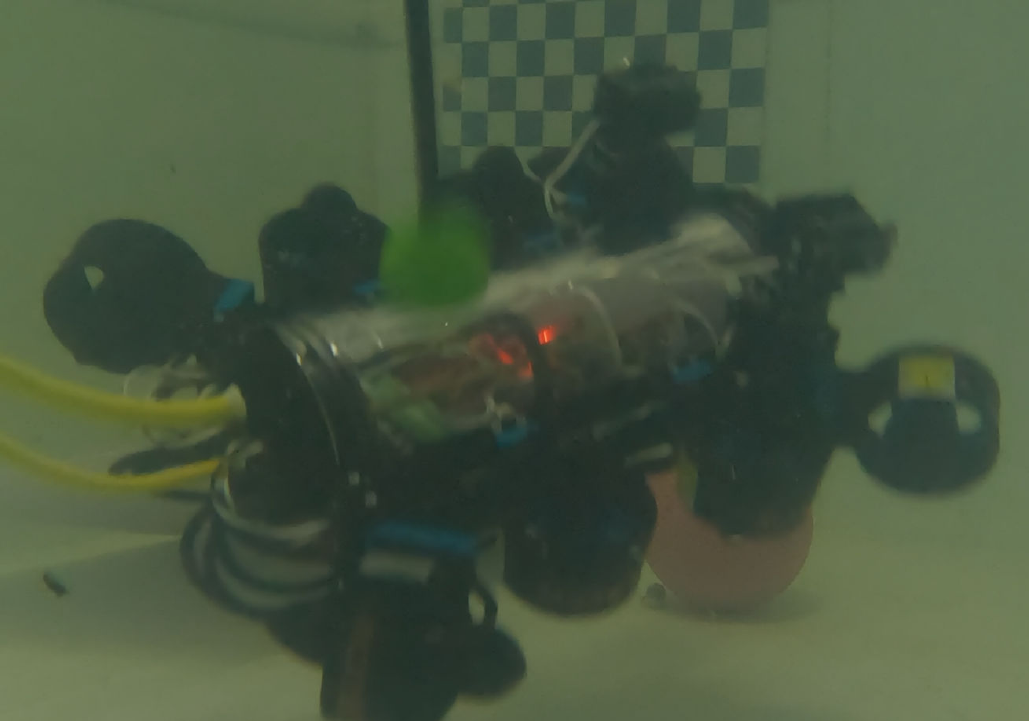}
    \caption{Forced instability}
    \vspace{-0.1in}
    \label{Fig:stick1}
  \end{subfigure}  \hfill 
  \begin{subfigure}[b]{0.49\linewidth} 
    \includegraphics[width=\linewidth]{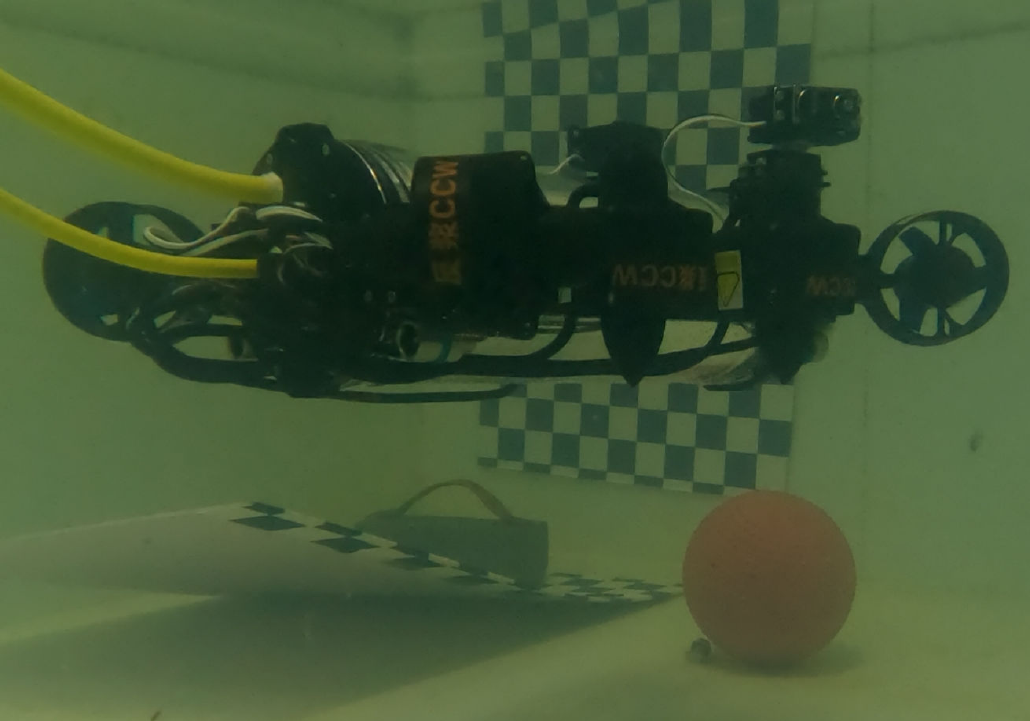}
    \caption{Back to horizontal}
    \vspace{-0.1in}
    \label{Fig:stick2}
  \end{subfigure}
  % 添加总体的标注
  \caption{Experiment for robust orientation control.}\label{Fig:stick_exp}
  \label{Fig:overall}
\end{figure}

The experiments in the previous subsection 
%could proceed normally, contingent upon 
depend on the maintenance of stable posture by each robot throughout the entire process of the localization algorithm.
To convincingly demonstrate the robustness of our platform, two experiments are designed in this subsection to verify the disturbance resilience  of the orientation control system. 

\begin{figure}[t]
  \centering\vspace{-0.1in}
  \includegraphics[width=\linewidth]{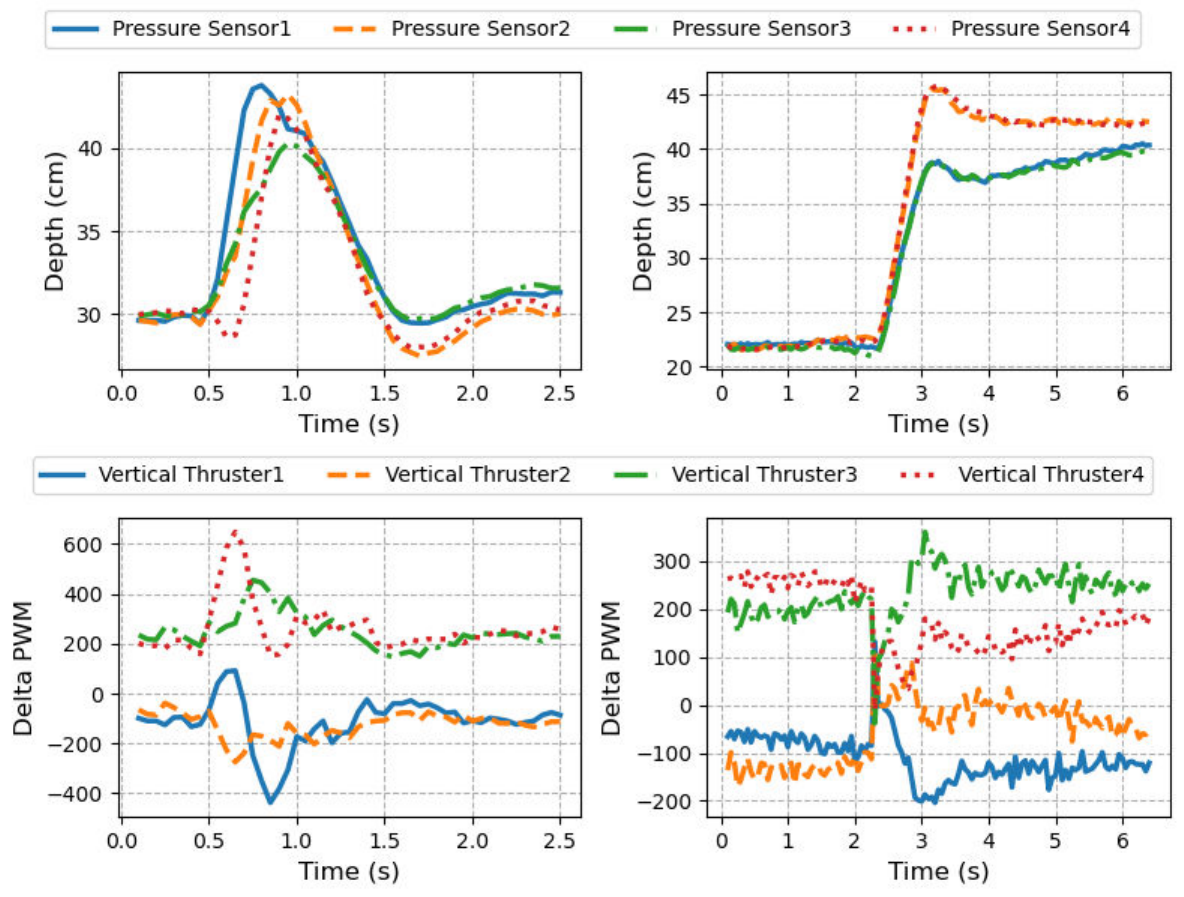}
  \vspace{-0.2in}
  \caption{Robust orientation control results.}
  \vspace{-0.1in}
\label{Fig:result}
%  \Description{A woman and a girl in white dresses sit in an open car.}

\end{figure}

The first experiment is external force disturbance experiment (Figure~\ref{Fig:stick_exp}). The robot suspending at the depth of 30cm is pushed by a stick (to simulate external interference). The second experiment is multi-depth suspension experiment, where  the robot descends from 22cm to 41cm and maintains horizontal orientation.
The data of water pressure sensors and the delta PWM input of the central thrusters are pensented in Figure~\ref{Fig:result}.

In the first experiment, despite being forcibly disrupted from a stable state by external disturbances, the system regain its original depth and horizontal orientation in \textbf{less than 2 seconds}. 
The second experiment achieves similar results. These two experiments validate the effectiveness and robustness of our orientation control framework.

%####################################################################
%—————————————————应用-------——————————
%####################################################################
    \vspace{-0.1in}
\subsection{Applications}
On the foundation of low-cost distributed localization, our platform can also be equipped with additional algorithms to realize a broader range of applications. 

\begin{figure}[t]
  \centering
  \begin{subfigure}{0.49\linewidth} 
    \includegraphics[width=\linewidth]{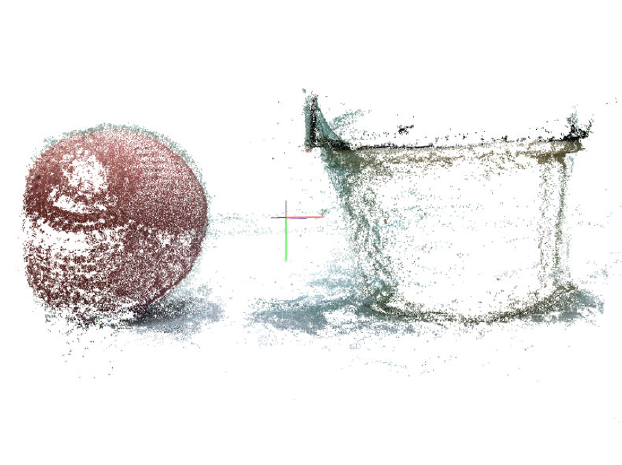} 
    \caption{Generated point clouds} 
    \vspace{-0.1in}
    \label{Fig:nerf_result}
  \end{subfigure}
\hfill
  \begin{subfigure}{0.49\linewidth} 
    \includegraphics[width=\linewidth]{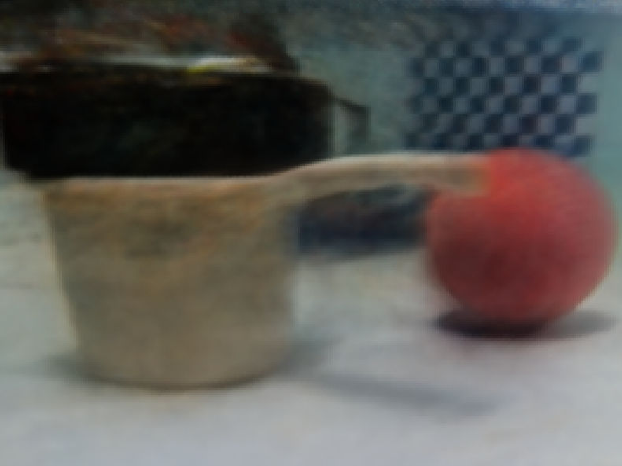} 
    \caption{Rendered image} 
    \vspace{-0.1in}
    \label{Fig:nerf}
  \end{subfigure}
  \caption{Application: underwater 3D reconstruction.}\vspace{-0.1in}\label{Fig:3d}
\end{figure}

\textbf{1) Underwater 3D reconstruction.} With the integration of the NeRF algorithm \cite{nerf}, we achieve distributed underwater 3D reconstruction in experimental environment (shown in Figure~\ref{Fig:env}). The results are depicted in Figure~\ref{Fig:3d}. 
%有了分布式定位算法，可以实现多台潜器同时对一个场景进行拍摄和重建，其间每个机器人都精确的知道需要重建的位置。这或有利于实现大型水下场景的重建，因为单个机器人能够拍摄范围有限，可以采用多个机器人来实现，我们的平台可以为这方面的应用提供底层的分布式定位支撑
Our distributed localization platform make it possible for multiple robots to simultaneously photograph and reconstruct a large-scale underwater scene, with each robot precisely obtaining the location that needs to be reconstructed.  
%This could be beneficial for the reconstruction of  large-scale underwater scenes, as the range that a single robot can photograph is limited, and multiple robots can be employed to achieve this. Our platform can provide foundational distributed localization support for such applications.

\begin{figure}[t]
  \begin{subfigure}{0.49\linewidth} 
    \includegraphics[width=\linewidth]{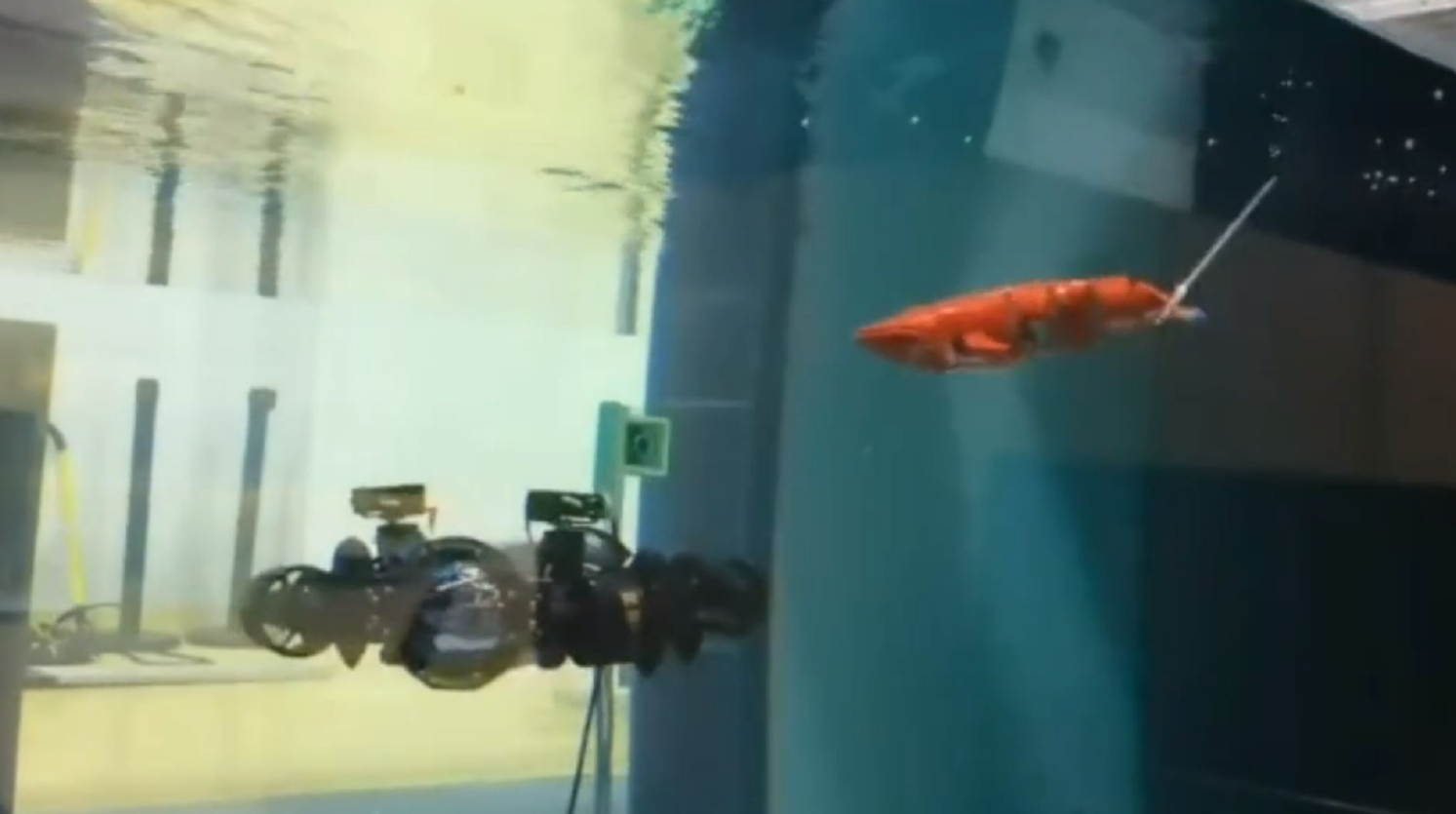} 
    \caption{External view} 
    \vspace{-0.1in}
    \label{Fig:fish1}
  \end{subfigure}
\hfill
  \begin{subfigure}{0.49\linewidth} 
    \includegraphics[width=\linewidth]{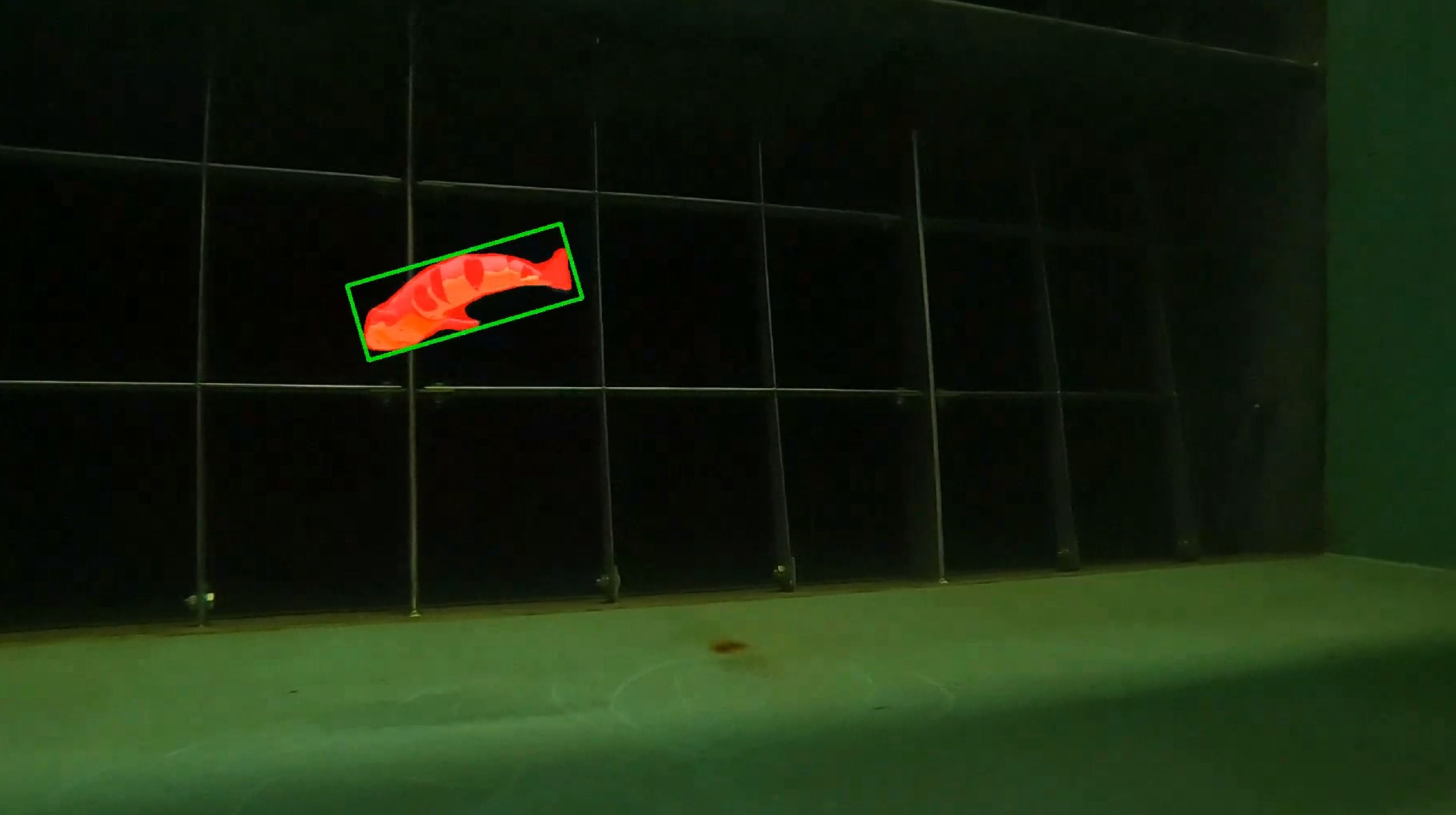} 
    \caption{Camera view} 
    \vspace{-0.1in}
    \label{Fig:fish2}
  \end{subfigure}

  \caption{Application: underwater creature tracking.}    \vspace{-0.1in}\label{Fig:fish}
\end{figure}

\textbf{2) Underwater creature tracking.}
 As shown Figure~\ref{Fig:fish}, by incorporating a target tracking algorithm \cite{nanotrack}, our platform is also capable of tracking aquatic organisms.


%我们的分布式定位算法能够获得目标生物的位置，有助于实现更稳定的跟踪。而且，由于分布式迭代算法的存在，多台机器人中只要有一台没有跟丢水下的目标生物，就能够通过分布式的迭代告诉其他机器人，让其他跟丢了的机器人重新跟踪上目标生物，大大提升了跟踪的鲁棒性。这将对于跟踪游动速度极快的生物，或者在复杂多遮挡环境下的生物追踪有着重要作用
Our multi-robot platform is capable of acquiring the location of the target organism, which aids in achieving more accurate tracking. Moreover, owing to our distributed protocol, as long as one robot in the platform does not lose sight of the underwater target, it can inform the others through the distributed iteration, allowing those that have lost track to reacquire the target organism, significantly enhancing the robustness of the tracking. This will be of significant importance for tracking organisms that move at extremely high speeds or those in complex, occlusion-rich environments.

%####################################################################
%—————————————————Conclusion-------------———————————
%####################################################################
\section{Conclusions}
In this paper, we presented Aucamp, an underwater multi-robot platform that utilize cost-effective monocular cameras to achieve distributed localization. 
%To the best of our knowledge, Aucamp is the first underwater robotic platform to systematically achieve low-cost yet effective camera-based perception, distributed underwater localization, and robust orientation control. 
%To solve the problem of ranging incapability of monocular cameras, w
%We proposed a Tenengrad-feature-based approach to accomplish depth estimation. An empirical formula for monocular localization was presented, coupled with detailed analysis of its error. 
We proposed a Tenengrad-feature-based approach to accomplish depth estimation, coupled with an empirical formula for monocular localization and detailed analysis of its error. 
Using distributed update protocol, we proposed an effective algorithm to accomplish global localization over the whole multi-robot platform. 
To support the distributed localization process of our underwater platform, we provided  rigorous dynamics model of the  multi-thruster robot in our platform and further designed the robust orientation control framework. Extensive experiments and application instances evaluated the effectiveness and robustness of out multi-robot platform. 
%The localization relative error was down to 0.4\%, and the robot only needs nearly 2 seconds to recover from forced instability. 
%We also employed several examples to demonstrate the broad potential for the application of our platform in various fields, including underwater 3D scene reconstruction and underwater creature tracking.

Future directions include designing a optimal cooperative illumination mechanism based on the imaging system of the proposed platform. The underwater domain, characterized by dimly lit conditions, can be advantageously illuminated by the supplemental lighting fixtures equipped on each robotic unit within our platform. The cooperative illumination capability may  greatly improve the viability and usability of camera-based platforms in underwater settings.

%\begin{acks}
%To Robert, for the bagels and explaining CMYK and color spaces.
%\end{acks}

%\bibliographystyle{unsrt}
\bibliographystyle{ACM-Reference-Format}
\bibliography{sample-base}

%\appendix

\end{document}